\documentclass[a4paper, conference]{IEEEtran}
\IEEEoverridecommandlockouts

\usepackage{cite}
\usepackage{float}
\usepackage{amsmath,amssymb,amsfonts}
\usepackage{algorithmic}
\usepackage{graphicx}
\usepackage{textcomp}
\usepackage{xcolor}
\usepackage{booktabs}
\usepackage{tabularx}
\usepackage{subcaption}
\usepackage[T1]{fontenc}
\usepackage{threeparttable}
\usepackage{hyperref}
\usepackage{siunitx}
\usepackage{flushend}

\def\BibTeX{{\rm B\kern-.05em{\sc i\kern-.025em b}\kern-.08em
    T\kern-.1667em\lower.7ex\hbox{E}\kern-.125emX}}
\begin{document}

\title{Design and Evaluation of Two Spherical Systems for Mobile 3D Mapping*\\

\thanks{*We acknowledge funding from the Elite Network Bavaria (ENB) 
for providing funds for the academic program ``Satellite Technology -- Advanced Space Systems''.}
}

\author{
\IEEEauthorblockN{
Marawan Khalil\IEEEauthorrefmark{2}\IEEEauthorrefmark{4},
Fabian Arzberger\IEEEauthorrefmark{2},
Andreas Nüchter\IEEEauthorrefmark{2}
}
\IEEEauthorblockA{\IEEEauthorrefmark{2}Computer Science XVII -- Robotics, Julius-Maximilians-Universität Würzburg, Germany}
\IEEEauthorblockA{\IEEEauthorrefmark{4}Faculty of MET -- CSEN Department, German University in Cairo, Egypt\\
marawansalah002@gmail.com, fabian.arzberger@uni-wuerzburg.de, andreas.nuechter@uni-wuerzburg.de}
}

\maketitle

\begin{abstract}
Spherical robots offer unique advantages for mapping applications in hazardous or confined environments, thanks to their protective shells and omnidirectional mobility. 
This work presents two complementary spherical mapping systems: a lightweight, non-actuated design and an actuated variant featuring internal pendulum-driven locomotion. 
Both systems are equipped with a Livox Mid-360 solid-state LiDAR sensor and run LiDAR-Inertial Odometry (LIO) algorithms on resource-constrained hardware. 
We assess the mapping accuracy of these systems by comparing the resulting 3D point-clouds from the LIO algorithms to a ground truth map.
The results indicate that the performance of state-of-the-art LIO algorithms deteriorates due to the high dynamic movement introduced by the spherical locomotion, leading to globally inconsistent maps and sometimes unrecoverable drift.
\end{abstract}

\begin{IEEEkeywords}
Spherical Robots, Mobile Mapping, Actuation Mechanisms, LiDAR, LiDAR-Inertial Odometry, SLAM.
\end{IEEEkeywords}

\section{Introduction}
\begin{figure}[t]
\centerline{\includegraphics[width=0.95\columnwidth]{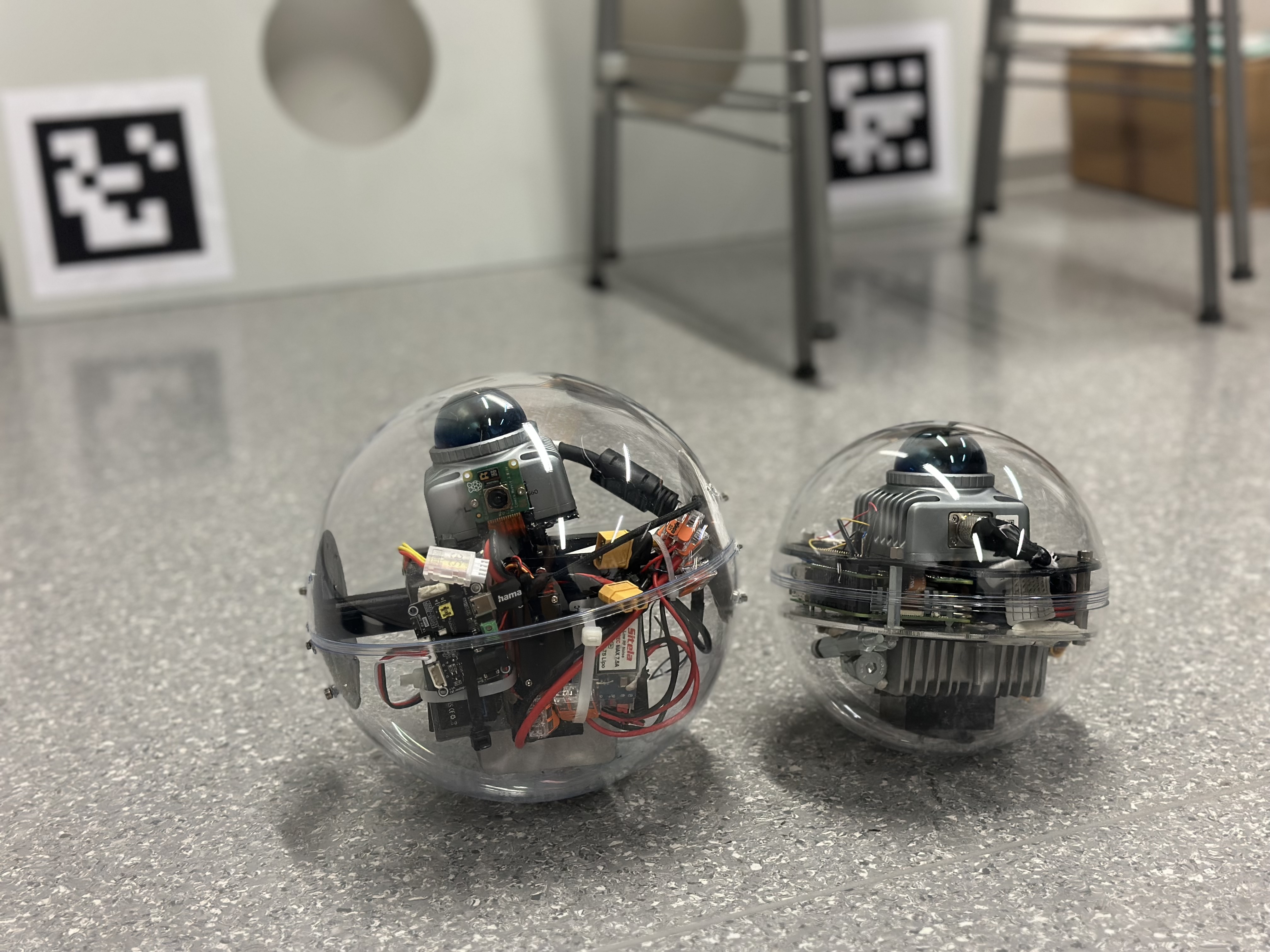}}
\caption{(Left:) \SI{20}{\centi\meter} diameter actuated sphere.
(Right:) \SI{16}{\centi\meter} diameter non-actuated sphere.
A video where both spheres are moving is available at \url{https://youtube.com/shorts/lxTF85HK-zY}.}
\vspace{-15pt}
\label{fig:twospheres}
\end{figure}

In recent decades, robotics has evolved from a niche discipline into a transformative technology impacting a wide range of industries, including manufacturing, healthcare, space exploration, and personal assistance. 
Among the diverse types of robotic systems, spherical robots have recently begun to attract increased attention from researchers. These robots represent a relatively unconventional design compared to more familiar, rotation-restricted systems such as UAVs, handheld devices, and wheeled vehicles. 
Unlike conventional robots, spherical robots offer several key advantages, including omnidirectional movement, enhanced maneuverability, and improved protection for internal components. 
These features make them well-suited for a variety of applications such as surveillance, and environmental monitoring in hazardous conditions. 
Their spherical design enables them to access environments that are difficult or impossible for other robotic systems to navigate, such as steep tunnels, and other confined or dangerous spaces.
The sealed outer shell of a spherical robot provides full protection against dust, hazardous chemicals, liquids, and external impacts. 
Recent research has focused on developing more robust motion mechanisms \cite{roboball,novelsphere,pendulum_sphere} and incorporating technologies like LiDAR for 3D mapping, further expanding their capabilities in complex and unpredictable environments\cite{Kalman_filter_sphere,DAEDALUS,sphere_Fabi_1}.
However, the rapid evolution of LiDAR–inertial algorithms and modern hardware platforms presents new opportunities for spherical SLAM, while the integration of actuation with mapping and localization remains largely unexplored.
To address this, we present and evaluate two custom-built spherical robots (see Fig.~\ref{fig:twospheres}) equipped with LiDAR-based SLAM systems.
Both systems integrate FAST-LIO2\cite{fastlio2}, FAST-LIVO2\cite{fastlivo2}, and DLIO\cite{dlio}, which are state-of-the-art LiDAR-inertial odometry algorithms.
In the next section, we review recent SLAM developments for spherical platforms, with emphasis on LiDAR-inertial methods, followed by advancements in spherical robot design.
In Section \ref{sec:hardwaredesing}, we provide a description of the hardware implementation of the two spherical robots. 
Subsequently, we describe the software integration, as well as the motion control mechanisms used for the actuated sphere. 
Finally, we provide a comprehensive evaluation of the mapping accuracy by comparing the maps of the proposed systems with a ground truth map obtained using high-precision terrestrial laser scanning (TLS). 

\section{State of the Art}

\subsection{Spherical SLAM}\label{AA}
Spherical Simultaneous Localization and Mapping (SLAM) is an emerging area within mobile robotics, offering promising solutions for robust mapping in constrained or hazardous environments. 
Unlike traditional platforms, spherical robots encapsulate their sensors within a protective shell and rely on rolling locomotion. 
This unique configuration introduces both opportunities and significant challenges for SLAM algorithms. 
One of the earliest spherical SLAM prototypes was L.U.N.A. \cite{luna}, which employed a 2D LiDAR and IMU inside a rolling spherical shell. 
The design validated the feasibility of spherical SLAM. 
It also featured actuation through internal flywheels, using an IBCOAM (Impulse by Conservation of Angular Momentum) mechanism.
A major milestone in this field was the DAEDALUS project \cite{DAEDALUS}, funded by the European Space Agency, which proposed a fully enclosed spherical robot for the autonomous exploration of lunar lava tubes. 
The robot was proposed to be equipped with LiDAR and internal actuators, and its design focused on resilience to lunar regolith and harsh environmental conditions.
A core challenge in spherical SLAM is the aggressive and off-centered rotation induced by rolling locomotion. 
These motions produce high angular velocities and dynamic behavior across all principal axes. 
This significantly degrades pose estimation accuracy and leads to error accumulation in the map. 
This problem is further compounded by the absence of magnetometer use—often intentionally excluded due to the unreliability of magnetic field data in planetary environments.
This leads to uncorrected yaw drift in IMU-based odometry, especially during prolonged navigation.
To address these issues, Arzberger et al. \cite{Kalman_filter_sphere,sphere_Fabi_1,DeltaFilter} introduced specialized filtering techniques for spherical systems. 
Their Delta Filter is a lightweight, real-time, multi-trajectory pose estimation method that fuses unreliable trajectories, such as those from IMUs and stereo visual-inertial odometry (VIO), into a more robust estimate, without requiring explicit sensor uncertainty modeling. 
The filter operates on pose changes ("deltas"), uses a probabilistic weighting scheme for translation estimation, and applies rotational interpolation via spherical linear interpolation (Slerp). 
A follow-up Kalman Filter design extended this approach by incorporating a covariance-aware model, enhancing pose estimation accuracy during rapid and complex motion.
While recent studies (e.g. \cite{Kalman_filter_sphere}) have compared their results with state-of-the-art SLAM systems such as DLIO\cite{dlio} and FAST-LIO2\cite{fastlio2}, the field continues to evolve rapidly. 
More recent SLAM algorithms (e.g., FAST-LIVO2 \cite{fastlivo2}), advanced LiDAR sensors like the MID-360 and RoboSense Airy, and modern single-board computers such as the Raspberry Pi 5 with PCIe support and Gigabit Ethernet are pushing the boundaries of what is possible. 
In this context, we introduce our first prototype, the non-actuated sphere, which leverages the processing power of the Raspberry Pi 5 16GB RAM model. 
This design enables real-time SLAM on a compact spherical platform, offering improved power efficiency, extended battery life, and reduced overall system cost.

\subsection{Spherical locomotion}\label{sec:state-of-the-art}
Spherical locomotion has attracted significant research interest in recent years, driven by the pursuit of optimal mobility mechanisms. 
Although spherical robots may appear mechanically simple, a wide variety of locomotion strategies have been developed—and continue to emerge—requiring sophisticated control systems to manage the complex dynamics of rolling locomotion and internal mass distribution.
One of the most well-known approaches is the Internal Driving Unit (IDU), or differential drive, used in commercial robots like the Sphero Bolt+ and BB-8. 
Akella et al.\cite{Sphero} analyzed BB-8’s internal wheel-driven system and demonstrated that effective control was achieved using only two actuators. 
However, they highlighted a key limitation: the robot’s geometry prevents controlled sliding and reduces effectiveness on inclined or uneven terrain.
In contrast, Zevering et al.~\cite{luna} introduced L.U.N.A., a spherical robot designed for autonomous 3D mapping in lunar caves. 
Rather than using wheels or rods, L.U.N.A. relies on internal flywheels to generate motion via the IBCOAM method. 
This approach enables a compact form factor and protects internal electronics, providing advantages particularly well suited to harsh and remote terrain. 
The robot demonstrated reliable motion on soft surfaces such as sand and rubber. 
However, limitations remain, including vibrational instability caused by unbalanced flywheels, reduced performance on inclined low-friction surfaces, and pose estimation errors due to unsynchronized sensor data.
In a following study, Zevering et al.~\cite{rod_sphere} proposed a rod-driven spherical robot, also targeting lunar cave exploration. 
This design uses external linear actuators to push against the environment to induce motion. 
While it improves adaptability to rugged terrain and sharp obstacles, it introduces new challenges, such as oscillatory behavior from fixed-speed actuators, limited effectiveness on slopes, and the need for higher-power actuation when traversing dusty or soft ground.
Beyond differential and flywheel-based designs, several researchers have explored pendulum-driven locomotion due to its mechanical simplicity, energy efficiency, and natural stability. 
Oevermann et al.~\cite{roboball}, Ren et al.~\cite{novelsphere}, and Kolbari et al.~\cite{pendulum_sphere} developed spherical robots that use an internal heavy pendulum as the primary driving mechanism. 
While their configurations differ in terms of shell design and target applications, they share a common reliance on pendulum-based locomotion and have demonstrated robust movement across a variety of terrains.
A notable example is RoboBall~\cite{roboball}, which features a novel soft pressurized shell and a two-degree-of-freedom internal pendulum. 
This robot successfully navigates gravel, grass, steep inclines, and even floats and maneuvers on water. 
However, the deformable shell introduces complex dynamic behavior. 
The researchers addressed this using a Linear Quadratic Regulator (LQR) for steering and a model-based proportional controller for driving. 
Ren et al. \cite{novelsphere} experimented with both robust servo LQR (RSLQR) and PID-based stabilization strategies to enhance motion control and responsiveness, whereas Kolbari et al. \cite{pendulum_sphere} adopted only PID control for stabilization. 
RoboBall’s experiments further revealed that internal pressure and shell deformation significantly affect dynamic behavior—especially due to the presence of a ``dead zone'' where balance control becomes unstable during motion.
Taken together, these pendulum-based locomotion strategies highlight the value of combining mechanical simplicity with robust control to enable adaptive movement in unpredictable environments. 
While a variety of innovative locomotion methods have been proposed for spherical robots, including flywheel-, rod-, and pendulum-driven systems, few have been evaluated in conjunction with advanced LiDAR-inertial odometry algorithms under real-world motion dynamics.
This paper addresses this gap by introducing our second prototype, a pendulum-driven spherical robot designed to achieve both robust locomotion and accurate, real-time mapping performance within the constraints of a compact platform.

\section{Hardware and Design}\label{sec:hardwaredesing}
This section describes the hardware design and structural implementation of two spherical robots developed for 3D mapping: the non-actuated sphere and the actuated sphere.

\subsection{Non-Actuated Sphere}
The non-actuated sphere was modeled using CAD software, drawing inspiration from the design by Arzberger et al.~\cite{Kalman_filter_sphere}.
Its structure features two flat discs stacked vertically and secured with pillar screws, maintaining a gap of approximately \SI{28}{\milli\meter} between them. 
To enhance structural integrity and safety, fillets were applied to the edges of the discs. 
Each disc has a uniform thickness of \SI{3}{\milli\meter}.
Two fabrication methods were used to produce the discs: 3D printing with PLA and laser cutting with acrylic. 
Table~\ref{tab:hardware_components_non_actuated} summarizes the components used in the non-actuated sphere and their locations within the sphere.
\begin{figure}
\centering
\begin{subfigure}{0.4\columnwidth}
    \centering
    \includegraphics[width=\textwidth]{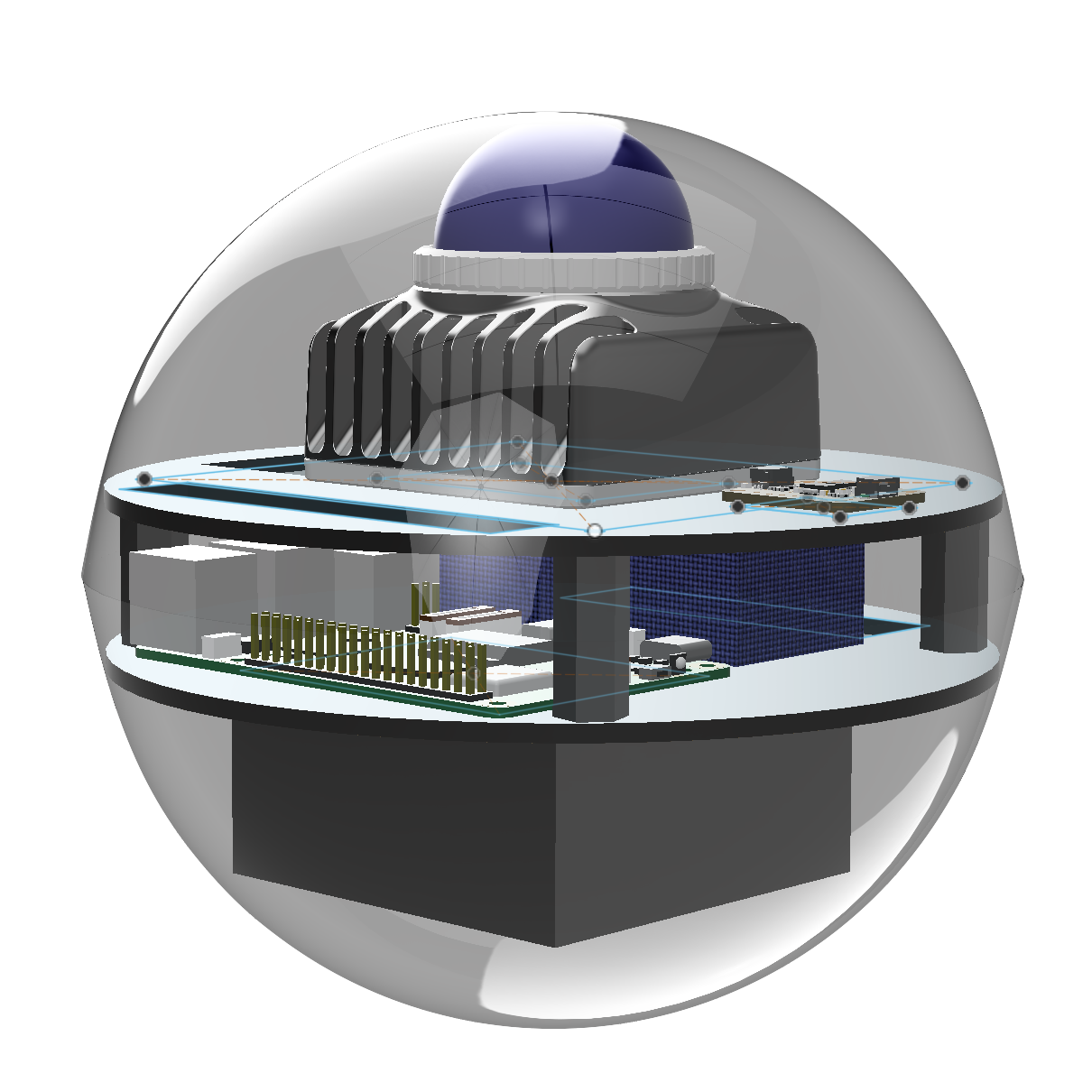}
    \caption{3D CAD model}
    \label{fig:cad-model}
\end{subfigure}
\hfill
\begin{subfigure}{0.4\columnwidth}
    \centering
    \includegraphics[width=\textwidth]{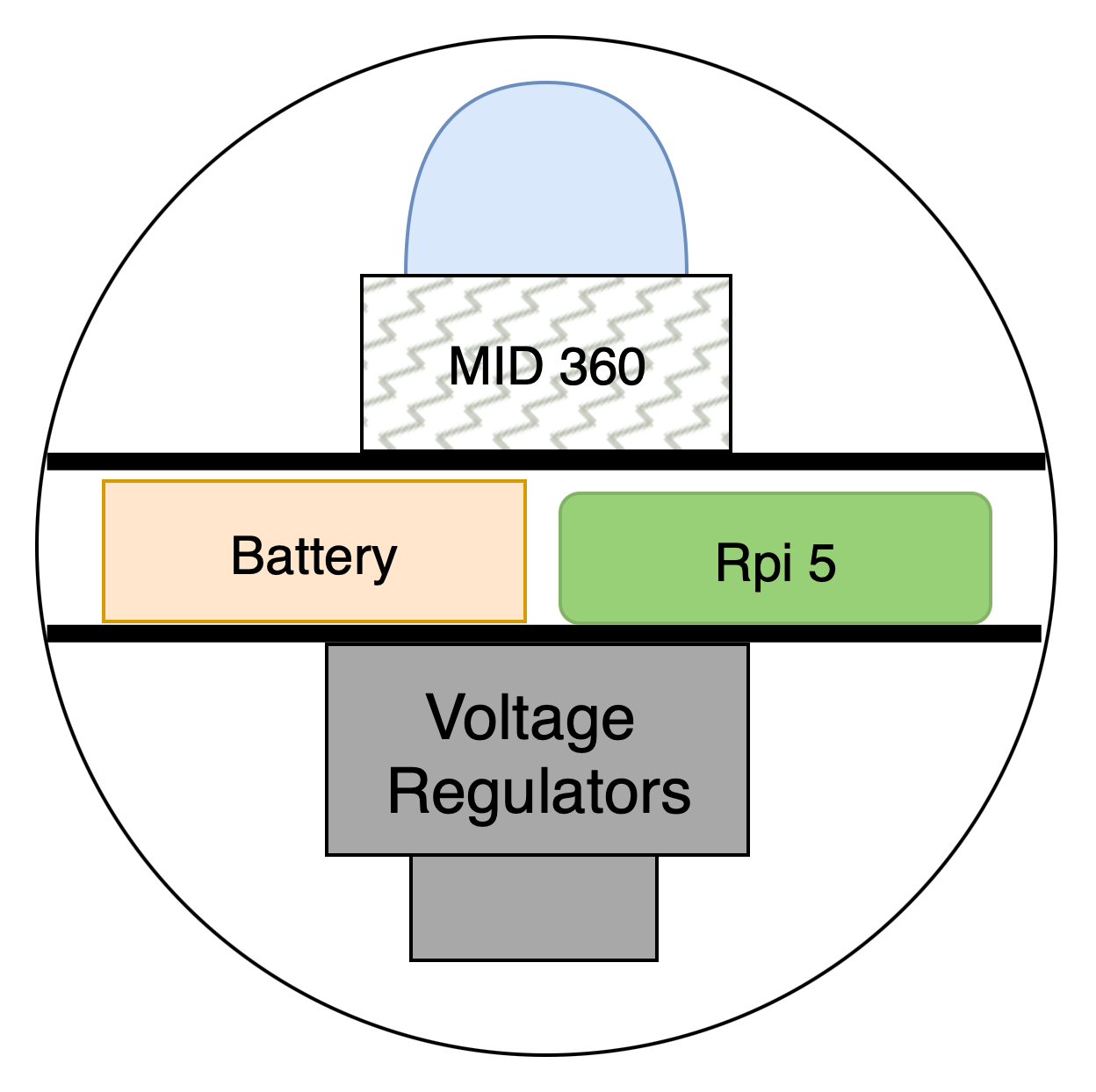}
    \caption{Simplified 2D Model}
    \label{fig:2d-model}
\end{subfigure}
\caption{Schematic model and design of the non-actuated sphere.
It must be pushed manually.}
\label{fig:cad-design1}
\end{figure}
\begin{table}
\centering
\caption{Hardware components and placement in the sphere}
\label{tab:hardware_components_non_actuated}
\begin{tabularx}{\linewidth}{@{}l X@{}}
\toprule
\textbf{Layer} & \textbf{Components} \\
\midrule
Top    & Livox Mid-360 LiDAR, BNO-085 IMU \\
Middle & Raspberry Pi 5 (16 GB, 256 GB SSD, cooling fan) \\
       & 2200 mAh 3S LiPo battery \\
Bottom & Voltage regulators \\
\bottomrule
\end{tabularx}
\vspace{-1em}
\end{table}
The Livox Mid-360 is connected to the Raspberry Pi via Ethernet, while the BNO-085 IMU communicates through the I2C interface. 
The sphere is designed to be lightweight, with a total weight of less than \SI{1}{\kilo\gram}, including the battery and all internal components. 
The sphere's diameter is \SI{16}{\centi\meter}. 
It is designed to be rolled manually by foot or hand.
To ensure stable and predictable rotation, metal weights were placed within the sphere to align its center of mass with its geometric center. 
A 3D CAD model of the assembled structure is shown in Fig.~\ref{fig:cad-design1}.

\subsection{Actuated Sphere}
\begin{figure}
\begin{subfigure}{0.47\columnwidth}
    \centering
    \includegraphics[width=\textwidth]{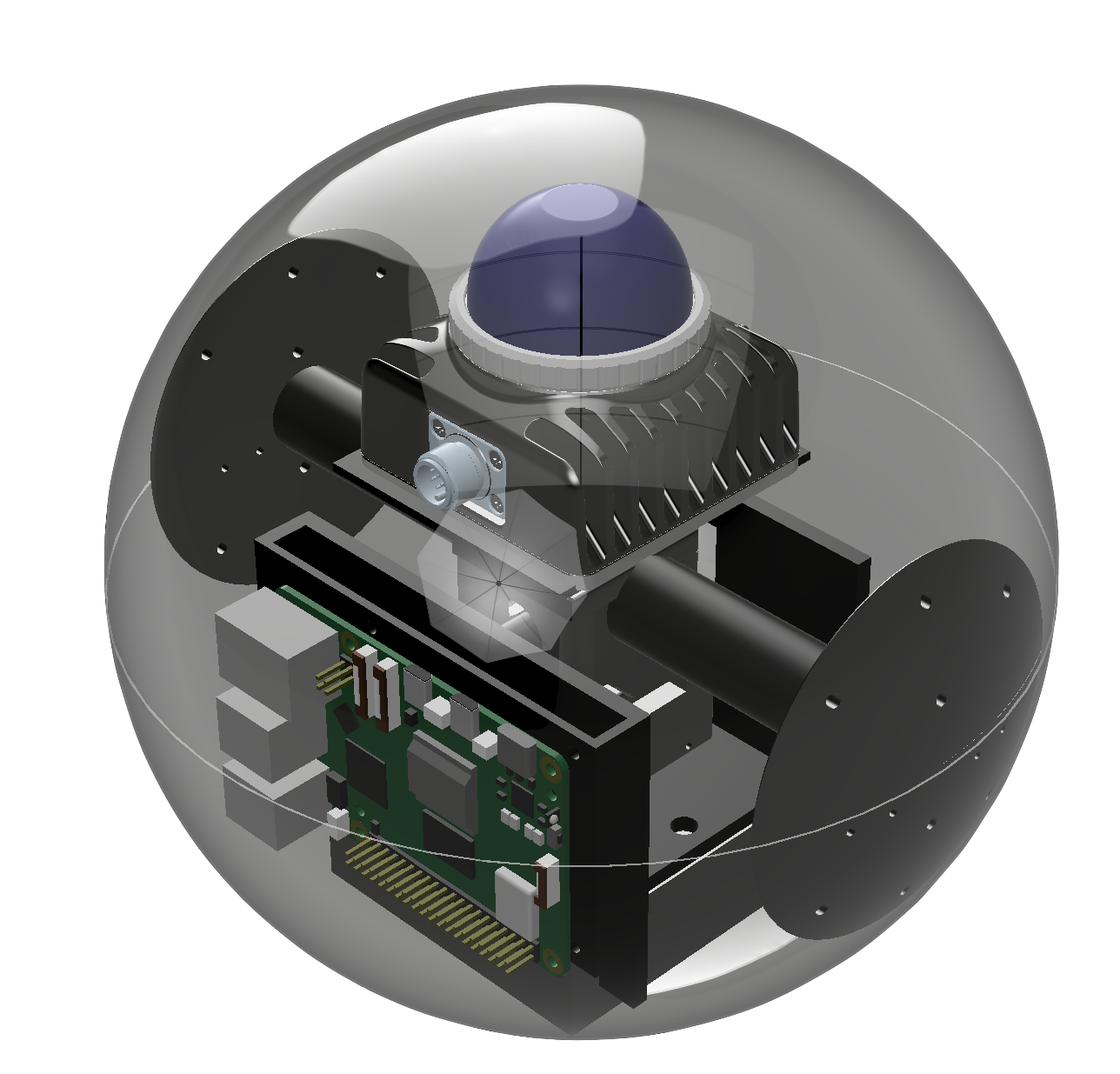}
    \caption{3D CAD model}
    \label{fig:cad-design2}
\end{subfigure}
\hfill
\begin{subfigure}{0.52\columnwidth}
    \centering
    \includegraphics[width=\textwidth]{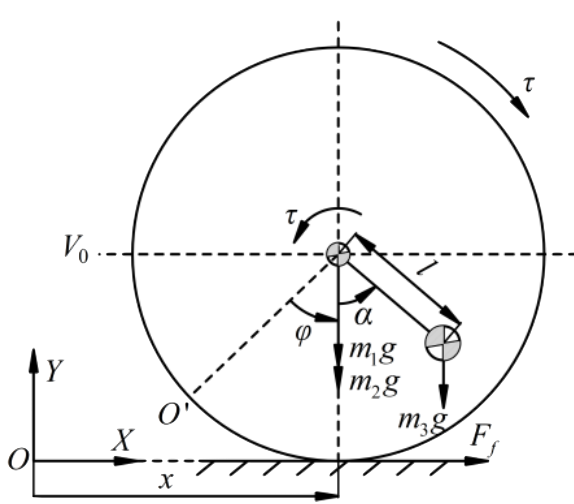}
    \caption{Pendulum model~\cite{Sphere_2D}}
    \label{fig:2d-model2}
\end{subfigure}
\caption{Schematic model and design of the actuated sphere.
The sphere uses a servo motor for actuation through pendulum-driven locomotion.
A second servo shifts the center of mass laterally by moving the battery to enable curved trajectories.}
\label{fig:cad-design2}
\end{figure}

The actuated sphere was designed using CAD software, with Fig.~\ref{fig:cad-design2} illustrating the final structure. 
As described in Section~\ref{sec:state-of-the-art}, the design was inspired by the pendulum-driven locomotion mechanisms presented in studies~\cite{roboball, novelsphere}.
Unlike the non-actuated version, this sphere features an internal actuation system. 
At its core, a Waveshare Smart Continuous Servo (model ST3215) is mounted to control forward and backward motion by dynamically shifting the internal center of mass in the direction of travel. 
A second PWM-controlled servo enables left and right movement via pendulum displacement along the lateral axis.
The structural components were fabricated using PLA filament and 3D-printed in multiple separate parts, which were then assembled into the complete sphere.
The system dynamics shown in Fig.~\ref{fig:2d-model2} involve key parameters including the robot’s horizontal position \( x \), the spinning angle of the shell \( \phi \), the swinging angle of the counter-weight mass \( \alpha \), and the input torque of the primary motor \( \tau \). 
Associated physical properties include the shell mass \( m_1 \), inner driver mass \( m_2 \), counter-weight mass \( m_3 \), gravitational acceleration \( g \), and the length of the connecting rod \( l \). 
The internal hardware components are organized according to their position within the sphere, as detailed in Table~\ref{tab:hardware_components_actuated}.
\begin{table}
\centering
\caption{Hardware components and placement in the sphere}
\label{tab:hardware_components_actuated}
\begin{tabularx}{\linewidth}{@{}l X@{}}
\toprule
\textbf{Position} & \textbf{Components} \\
\midrule
Center & Waveshare Smart Continuous Servo (model ST3215) \\
       & Forward/backward actuation \\
       & PWM Servo – left/right pendulum control \\
Rear   & 2200 mAh 3S LiPo battery \\
       & 5V 5A voltage regulator (for Raspberry Pi 5) \\
       & 6V UBEC (for servo motors) \\
Pendulum Module & 12V 20A voltage regulator – powers the entire system; positioned near the shell for stability \\
Front  & Raspberry Pi 5 (16 GB model) \\
Top    & Livox Mid-360 LiDAR \\
       & Pi Camera V3 (12 MP) \\
\bottomrule
\end{tabularx}
\vspace{-4mm}
\end{table}

\begin{figure*}
    \centering
    \includegraphics[width=1\linewidth]{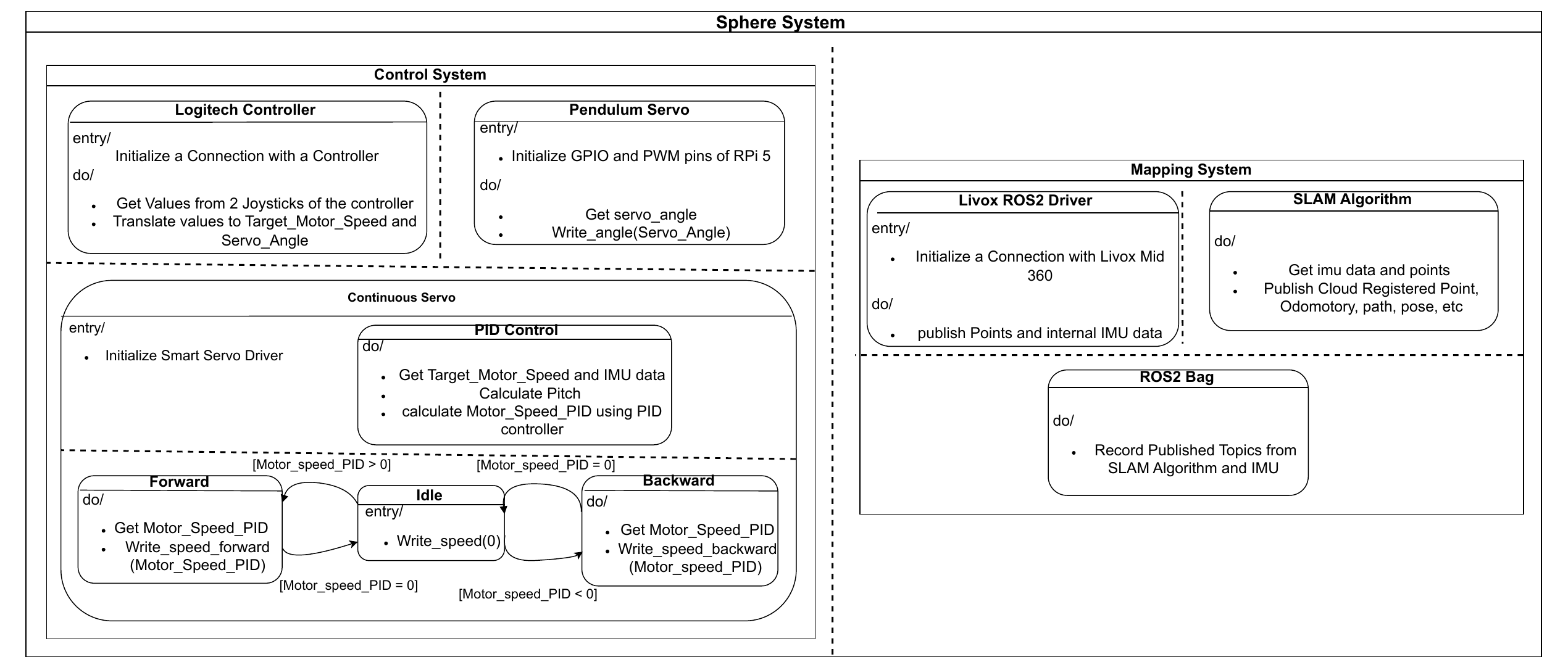} 
    \caption{State chart of the Sphere System showing the control and mapping subsystems.}
    \label{fig:sphere_system}
\end{figure*}

\section{Software and Control System}
As shown in the system state diagram in Fig.~\ref{fig:sphere_system}, both the actuated and non-actuated spheres follow the same state transitions, with the exception of the control subsystem, which is unique to the actuated variant. 
All software components are deployed on Ubuntu 24.04.2 LTS ARM and ROS2 Jazzy.

\subsection{Mapping System}
Both spheres incorporate a real-time mapping system. 
The following LiDAR-Inertial Odometry (LIO) frameworks were evaluated and integrated:
\begin{itemize}
    \item \textbf{FAST-LIO2}
    \item \textbf{DLIO}
    \item \textbf{FAST-LIVO2} (used in LIO-only mode)
\end{itemize}
Some packages needed modification to meet the constraints and processing capabilities of the Raspberry Pi 5 and ROS2 Jazzy.
The source code is available in our GitHub repository~\cite{githubsphere}.
All mapping computations are performed onboard in real-time.

\subsection{Actuator System (Actuated Sphere Only)}
The actuated sphere features an internal movement control system, operated using a Logitech F710 controller. 
The controller's two joysticks are mapped to distinct control tasks:
\begin{itemize}
    \item \textbf{Servo control:} One joystick provides input to the continuous rotation servo. 
    These inputs are scaled and passed through a discrete Proportional-Integral-Derivative (PID) controller to regulate pitch by minimizing the error between the desired target angle and the current angle measured by the IMU.
    \item \textbf{Mass shifting:} The second joystick adjusts the angle of an internal weight to shift the center of mass left or right, enabling directional control of the sphere's rolling behavior.
\end{itemize}
This control strategy allows for fine-grained movement and stabilization, combining feedback-based servo control with physical mass displacement for maneuvering.

\section{Results and Evaluation}
In this section, we present the evaluation and results of both spherical robots in terms of their mapping accuracy and stability.
All experiments were conducted in a controlled indoor environment within the Computer Science building at the University of Würzburg.

\subsection{Evaluation Setup}
\begin{figure}[t]
\centering
\begin{subfigure}{0.495\columnwidth}
        \centering
        \includegraphics[width=\textwidth]{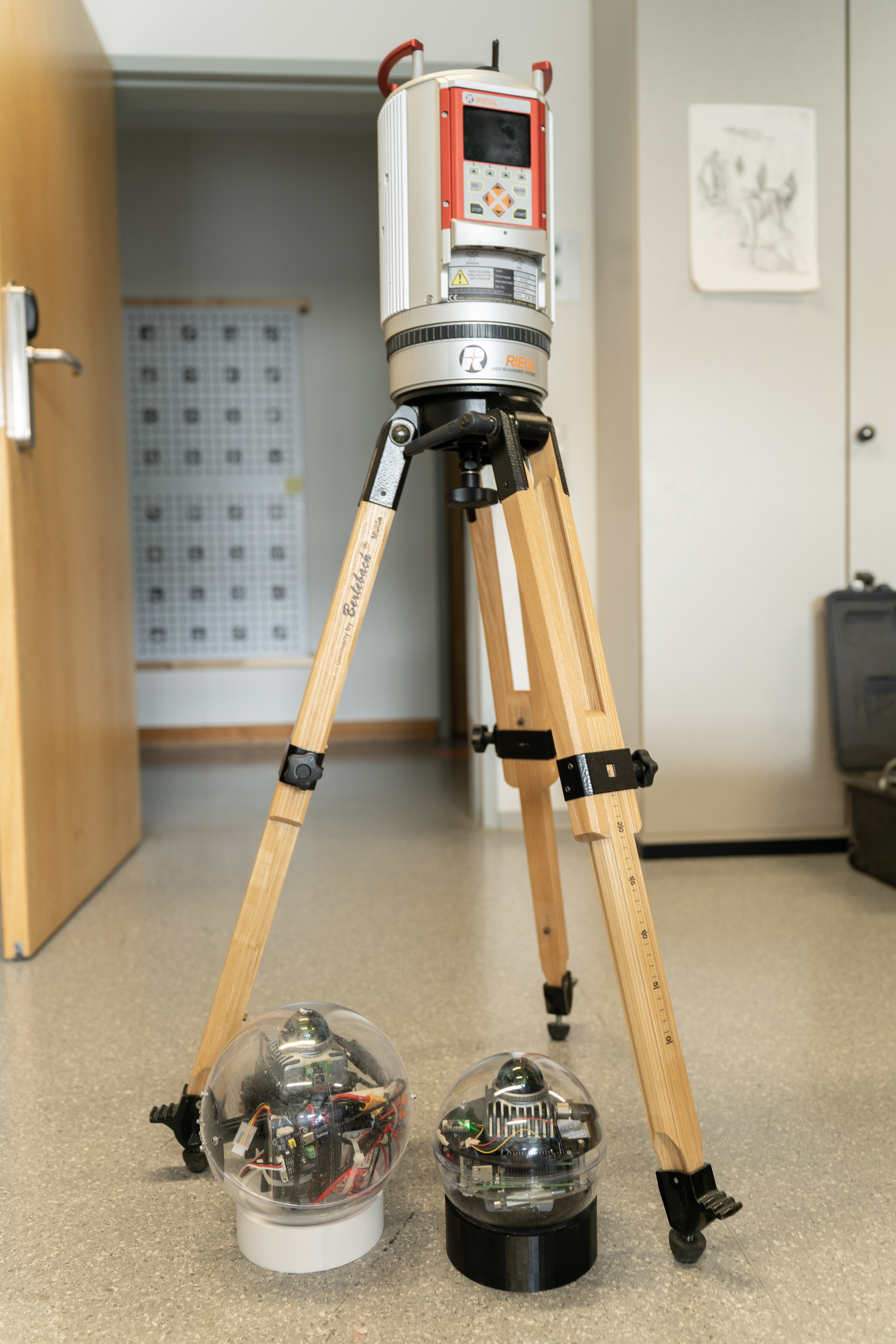}
        \caption{Riegl VZ-400}
        \label{fig:riegl}
\end{subfigure}
\hfill
\begin{subfigure}{0.49\columnwidth}
        \centering
        \includegraphics[width=\textwidth]{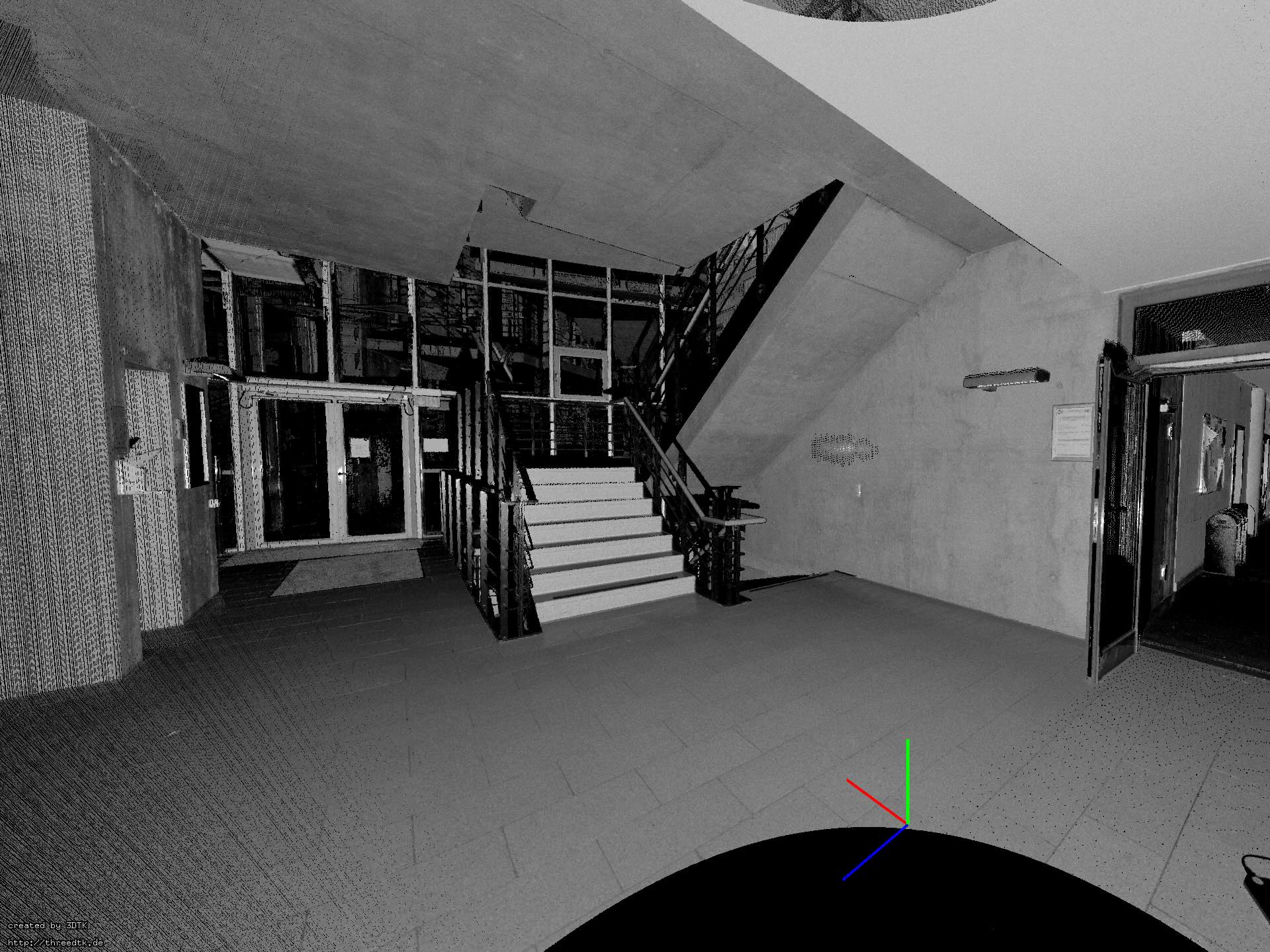}\vspace{.5mm}
        \includegraphics[width=\textwidth]{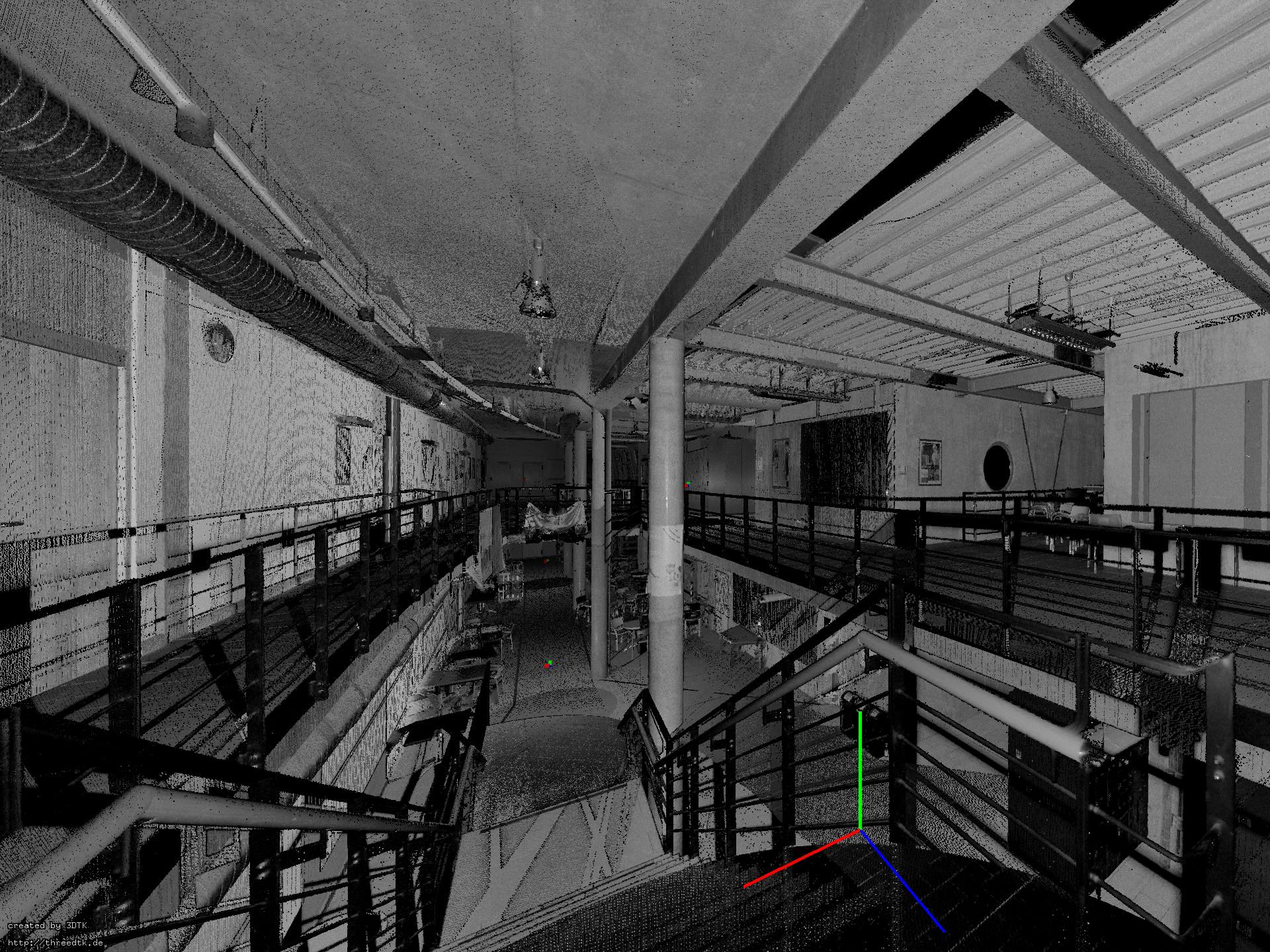}
        \caption{Ground truth point-clouds}
        \label{fig:sphere_on_the_move}
\end{subfigure}
\caption{Evaluation setup, showing the Riegl VZ-400 terrestrial laser scanner (TLS) and the resulting point-cloud used as ground truth in the evaluation.}\vspace{-3mm}
\end{figure}

The evaluation consisted of mapping selected indoor areas—specifically, a corridor, a hall, and the upper floor—using both the non-actuated and actuated spherical robots. 
For the non-actuated sphere, the robot was manually moved by hand and by foot. 
Each mapping run used one of three SLAM algorithms: FAST-LIO2, DLIO, and FAST-LIVO2 (LIO mode). 
The experiment was repeated three times—once for each algorithm—under the same environmental conditions. 
The same procedure was carried out with the actuated sphere, with the key difference being that it was moved using a controller rather than manually. 
Again, each of the three SLAM algorithms was executed along similar paths within the same locations. 
We used 3DTK~\cite{3dtk} to process the resulting point-clouds, which were evaluated by comparing them against ground truth data obtained using a Riegl VZ-400 terrestrial laser scanner (TLS) as shown in Fig.~\ref{fig:riegl}.
\begin{figure}[t]
\centering
\begin{subfigure}{0.492\columnwidth}
        \centering
        \includegraphics[width=\textwidth]{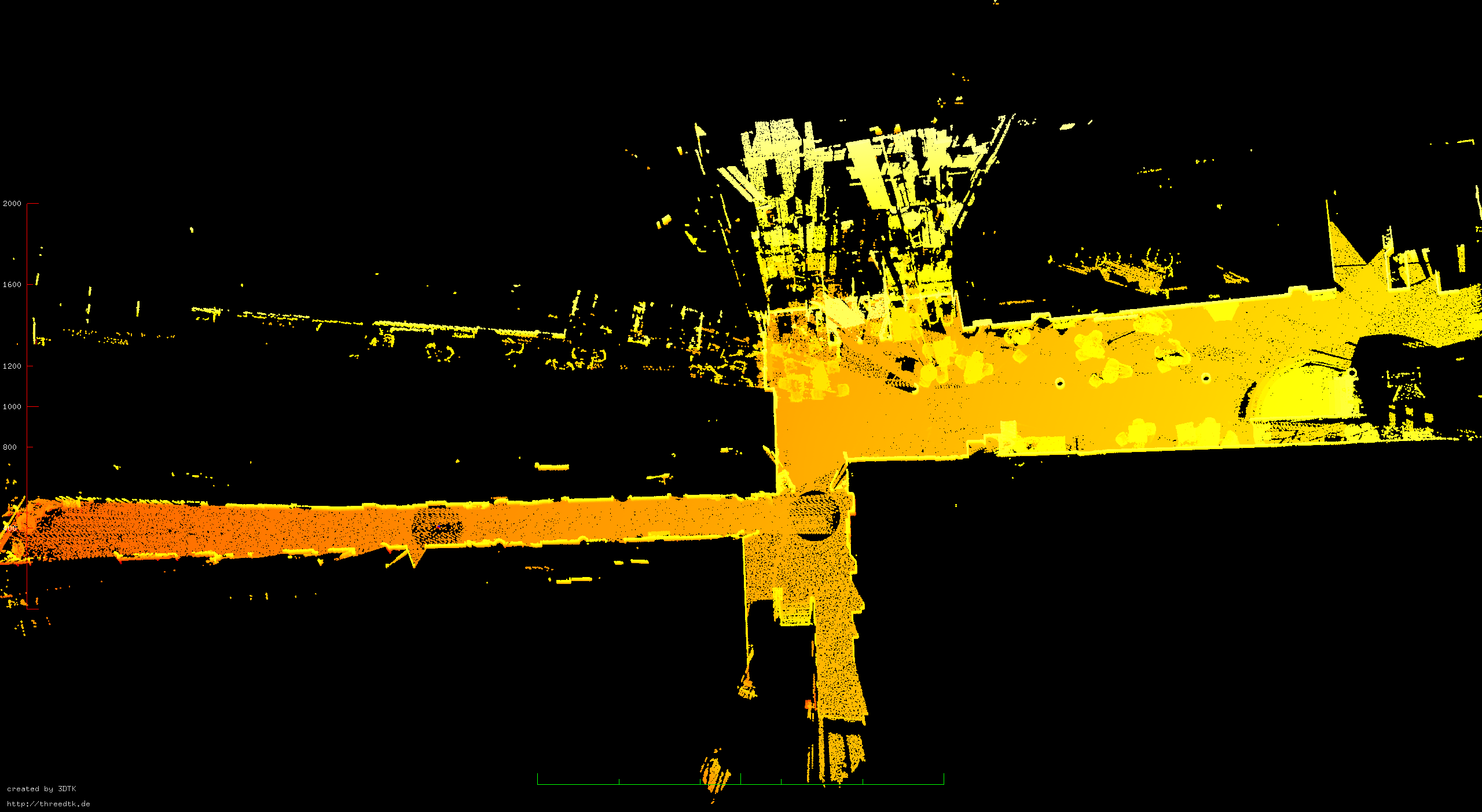}
        \caption{Riegl Map Result}
        \label{fig:riegl_top}\end{subfigure}
\hfill
\begin{subfigure}{0.492\columnwidth}
        \centering
        \includegraphics[width=\textwidth]{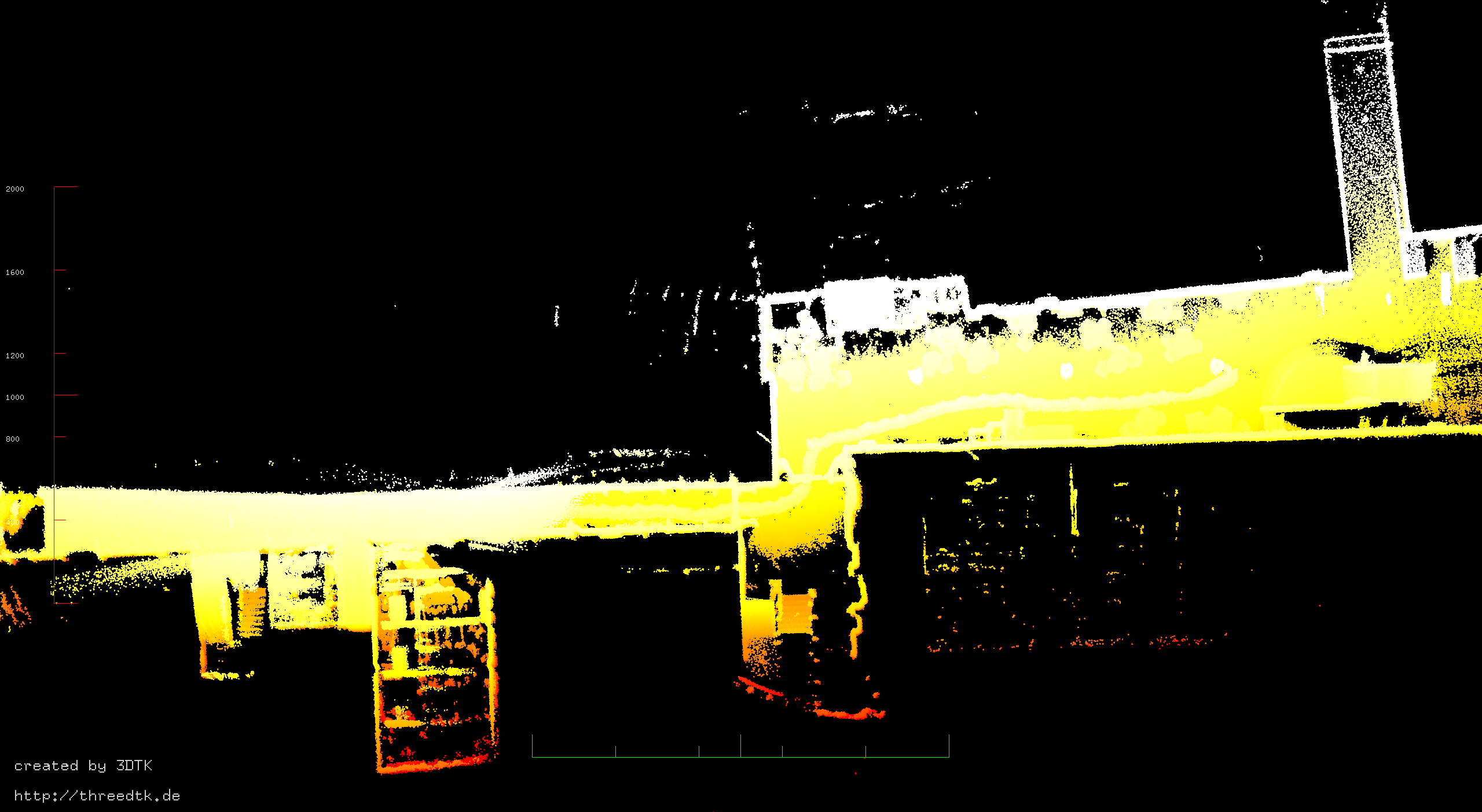}
        \caption{Actuated FAST-LIVO2}
        \label{fig:act_livo_top}
\end{subfigure}\vspace{2mm}
\begin{subfigure}{0.492\columnwidth}
        \centering
        \includegraphics[width=\textwidth]{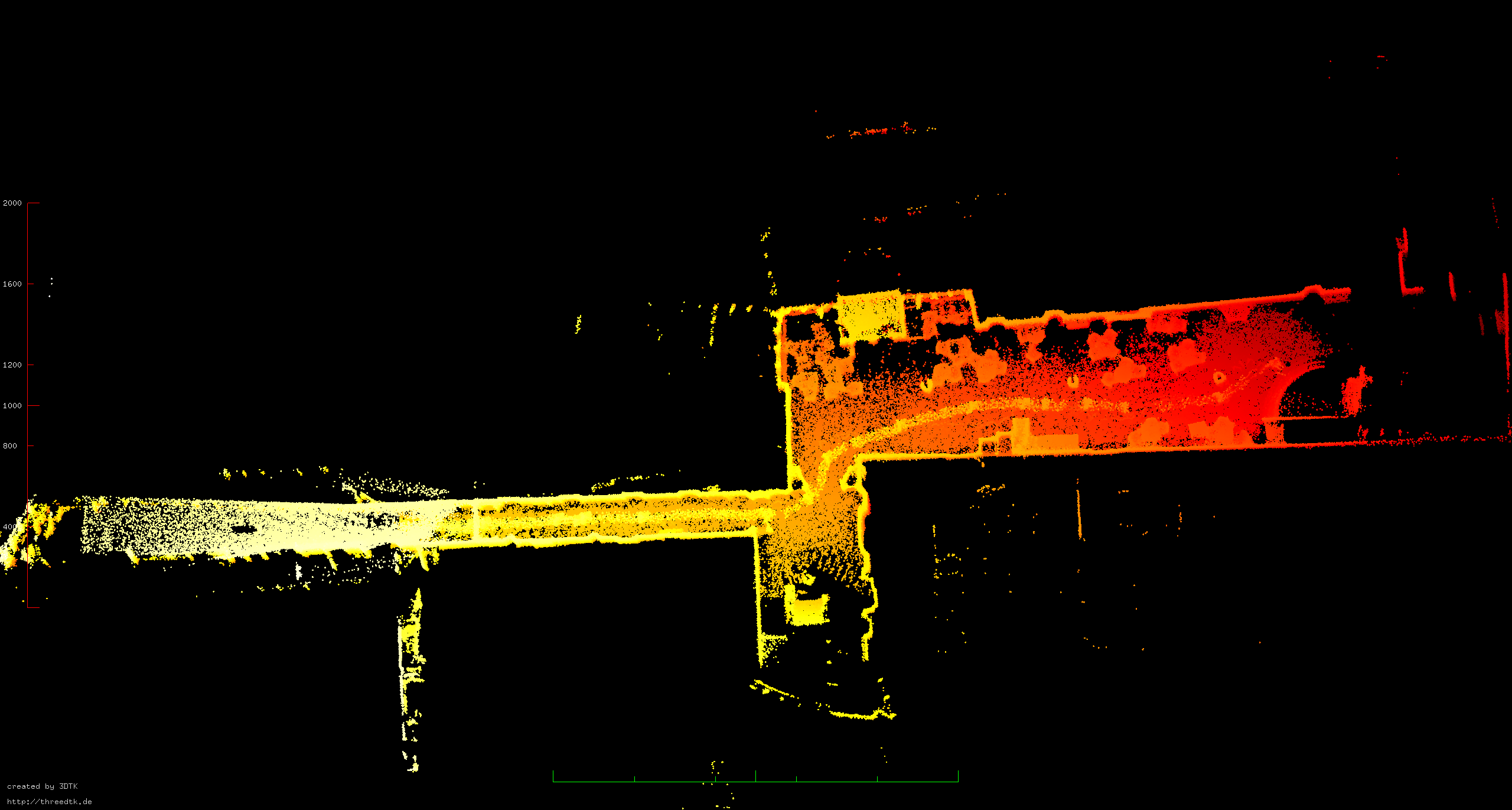}
        \caption{Actuated FAST-LIO2}
        \label{fig:act_fast_lio_top}\end{subfigure}
\hfill
\begin{subfigure}{0.492\columnwidth}
        \centering
        \includegraphics[width=\textwidth]{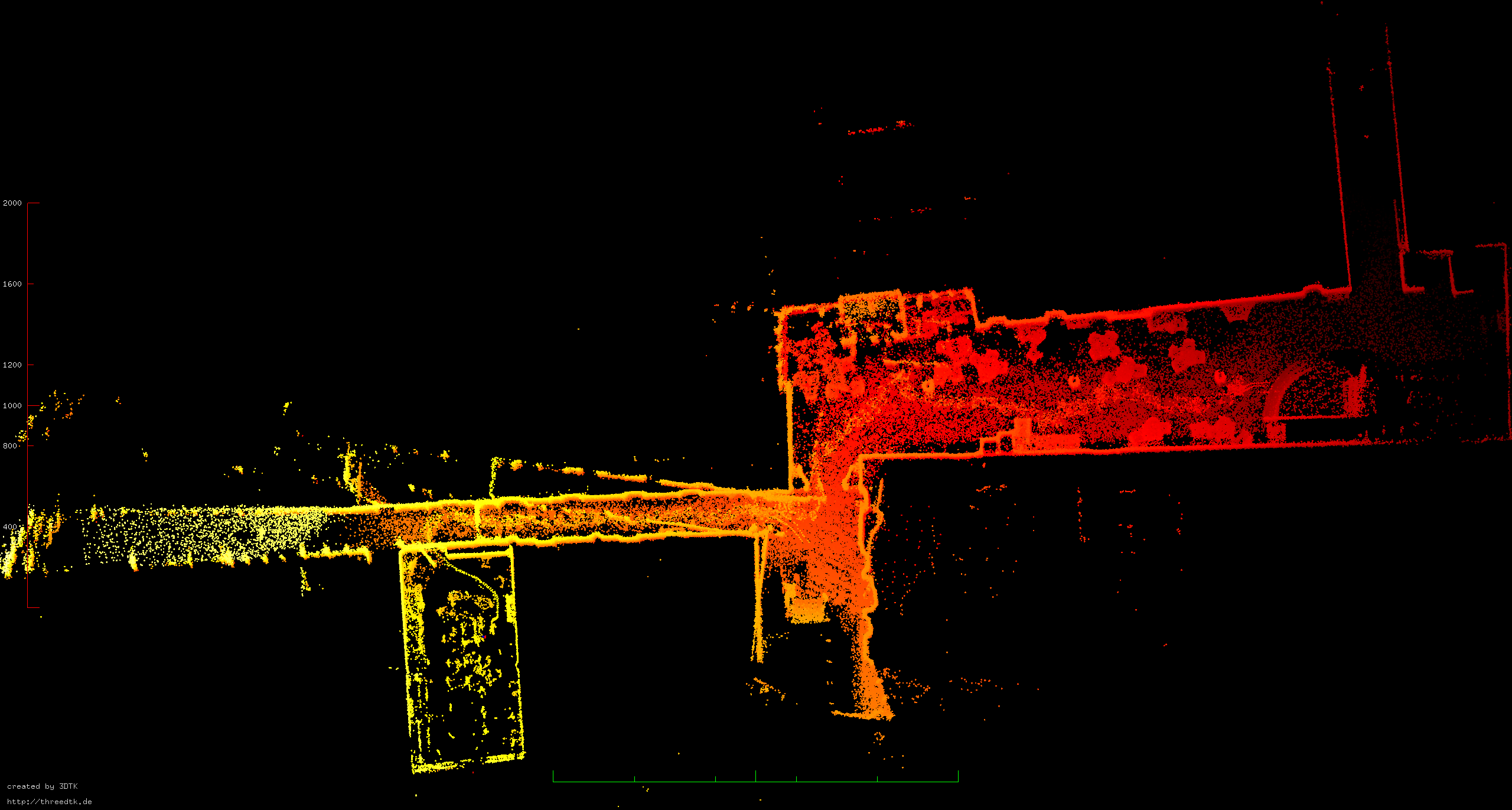}
        \caption{Non-actuated FAST-LIO2}
        \label{fig:non_act_fast_lio_top}
\end{subfigure}
\caption{Bird's-eye view of a cross section of resulting point-clouds (color indicates height where red means higher).}\vspace{-1mm}
\end{figure}

The error metric used was the point-cloud root mean square error (RMSE).
\begin{equation}
    RMSE = \sqrt{\frac{1}{N} \sum_{i=1}^{N} |p_i - q_i|^2}
\end{equation}
where \( p_i \) and \( q_i \) are the corresponding points in the model and data point-clouds, and \( N \) is the number of corresponding points. 
The RMSE measures the average distance between corresponding points in the two point clouds, providing an indication of the mapping accuracy.


\subsection{Evaluation Results}\label{AAA}
\subsubsection{Mapping Accuracy}
We evaluate the point-cloud error for each mapping run using the three SLAM algorithms. 
Table~\ref{tab:point_cloud_error} summarizes the results.
\begin{table*}
\centering
\begin{threeparttable}
\caption{Statistical analysis of point-cloud mapping accuracy.}
\label{tab:point_cloud_error}
\begin{tabular}{l|cc|cc|cc|cc|cc|cc}
\toprule
& \multicolumn{2}{c|}{\textbf{Points}} & \multicolumn{2}{c|}{\textbf{Mean$^1$}} & \multicolumn{2}{c|}{\textbf{Std Dev}} & \multicolumn{2}{c|}{\textbf{RMSE$^1$}} & \multicolumn{2}{c|}{\textbf{P95$^2$}} & \multicolumn{2}{c}{\textbf{P90$^2$}} \\
& \multicolumn{2}{c|}{\textbf{($\times 10^6$)}} & \multicolumn{2}{c|}{\textbf{(cm)}} & \multicolumn{2}{c|}{\textbf{(cm)}} & \multicolumn{2}{c|}{\textbf{(cm)}} & \multicolumn{2}{c|}{\textbf{(cm)}} & \multicolumn{2}{c}{\textbf{(cm)}} \\
\cmidrule(r){2-3} \cmidrule(lr){4-5} \cmidrule(lr){6-7} \cmidrule(lr){8-9} \cmidrule(l){10-11} \cmidrule(l){12-13}
\textbf{Algorithm} & \textbf{Non-act.} & \textbf{Act.} & \textbf{Non-act.} & \textbf{Act.} & \textbf{Non-act.} & \textbf{Act.} & \textbf{Non-act.} & \textbf{Act.} & \textbf{Non-act.} & \textbf{Act.} & \textbf{Non-act.} & \textbf{Act.} \\
\midrule
FAST-LIO2 & 1.99 & 4.43 & \bf{9.60} & 10.73 & 8.91 & 9.96 & \bf{13.09} & 14.64 & \bf{26.76} & 30.47 & \bf{18.58} & 22.57 \\
FAST-LIVO2(LIO Mode) & 16.36 & 46.39 & \bf{12.05} & 12.93 & 11.33 & 12.58 & \bf{16.54} & 18.04 & \bf{37.33} & 41.04 & \bf{26.48} & 31.63\\
DLIO & - & 88.01 & - & \bf{13.71} & - & 13.26 & - & \bf{19.08} & - & \bf{42.36} & - & \bf{34.70}\\
\bottomrule
\end{tabular}
\begin{tablenotes}
    \item (-) denotes that the algorithm failed.
    \item[1] Refers to the point-to-point errors to ground truth.
    \item[2] Refers to the percent of points having point-to-point errors to ground truth smaller than the given value. 
\end{tablenotes}
\end{threeparttable}
\end{table*}
The statistical analysis reveals significant differences in mapping performance across configurations. 
The non-actuated sphere with FAST-LIO2 achieved the lowest mean error (\SI{9.60}{\centi\meter}) and RMSE (\SI{13.09}{\centi\meter}), indicating superior accuracy. 
This is unexpected since the motion of the non-actuated sphere is governed by larger rotations around all principal axes and more aggressive dynamics.
We attribute the overall better mapping performance of the non-actuated sphere to its smaller shell radius and missing locomotion mechanism structure.
Thus, we were able to place the laser scanner closer to the center of the sphere.

\subsubsection{Drift and Bending}
\begin{figure}
\centering
\begin{subfigure}{0.49\columnwidth}
    \centering
    \includegraphics[width=\textwidth]{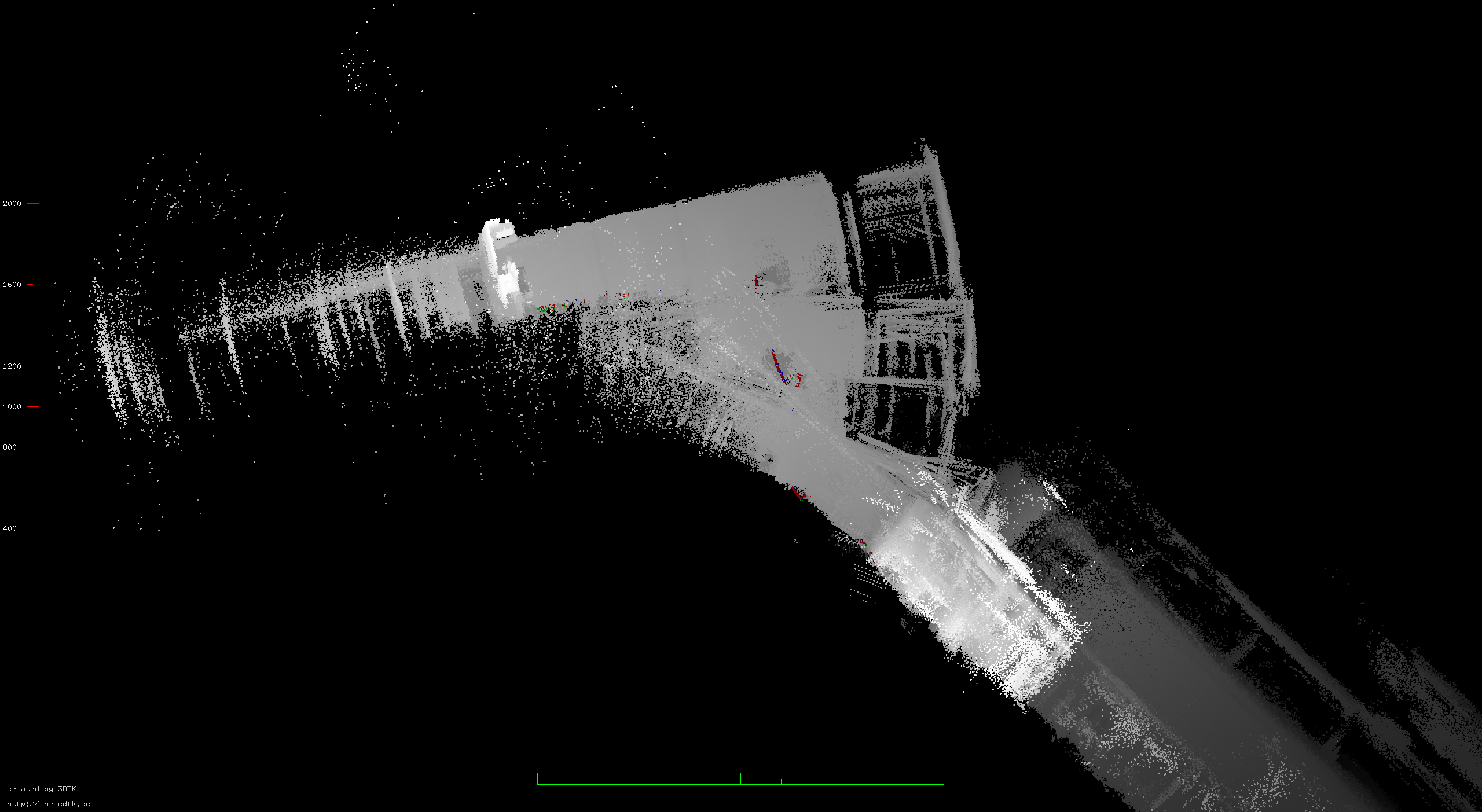}
    \caption{DLIO point-cloud}\label{fig:dlio_drift}
\end{subfigure}
\begin{subfigure}{0.49\columnwidth}
    \centering
    \includegraphics[width=\textwidth]{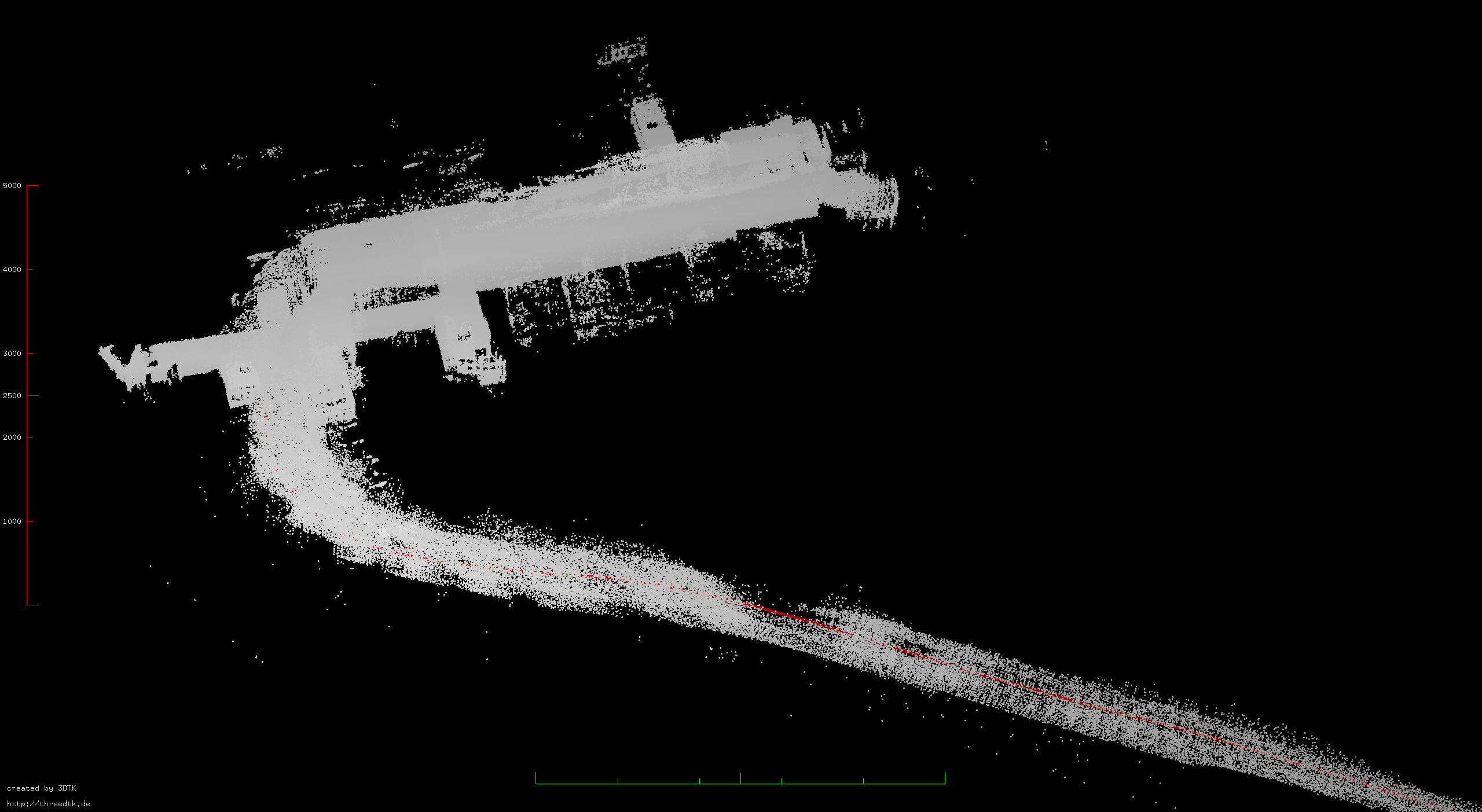}
    \caption{FAST-LIO2 point-cloud}\label{fig:lio_drift}
\end{subfigure}
\caption{Example failed cases from the non-actuated sphere where the LIO algorithm could not recover due to fast angular motion.
This results in huge drift and inconsistent mapping.}\vspace{-3mm}
\label{fig:drift}
\end{figure}

\begin{figure}
\centering
\begin{subfigure}{0.49\columnwidth}
    \centering
    \includegraphics[width=\textwidth]{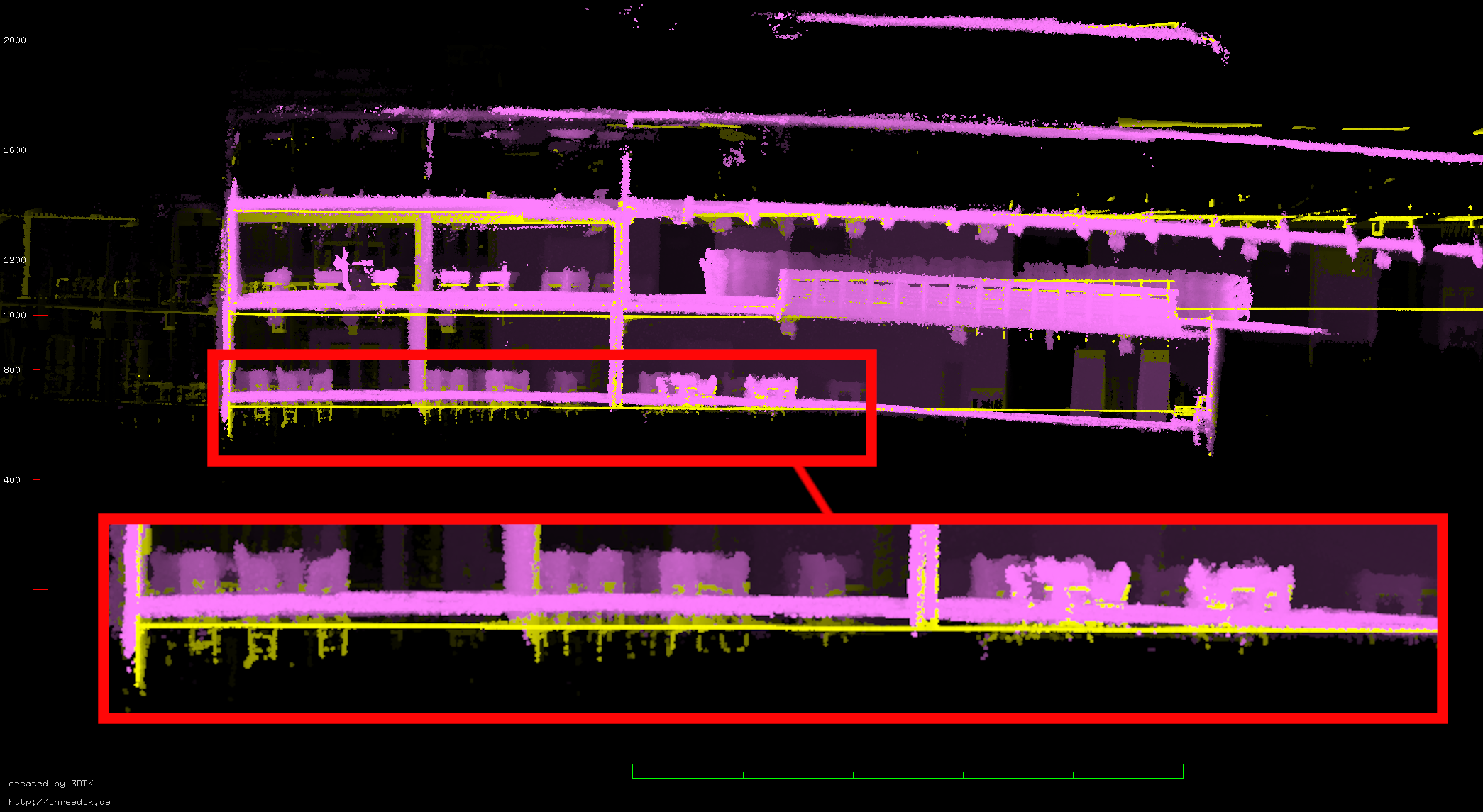}
    \caption{DLIO point-cloud}\label{fig:dlio_bending}
    \end{subfigure}
\begin{subfigure}{0.49\columnwidth}
    \centering
    \includegraphics[width=\textwidth]{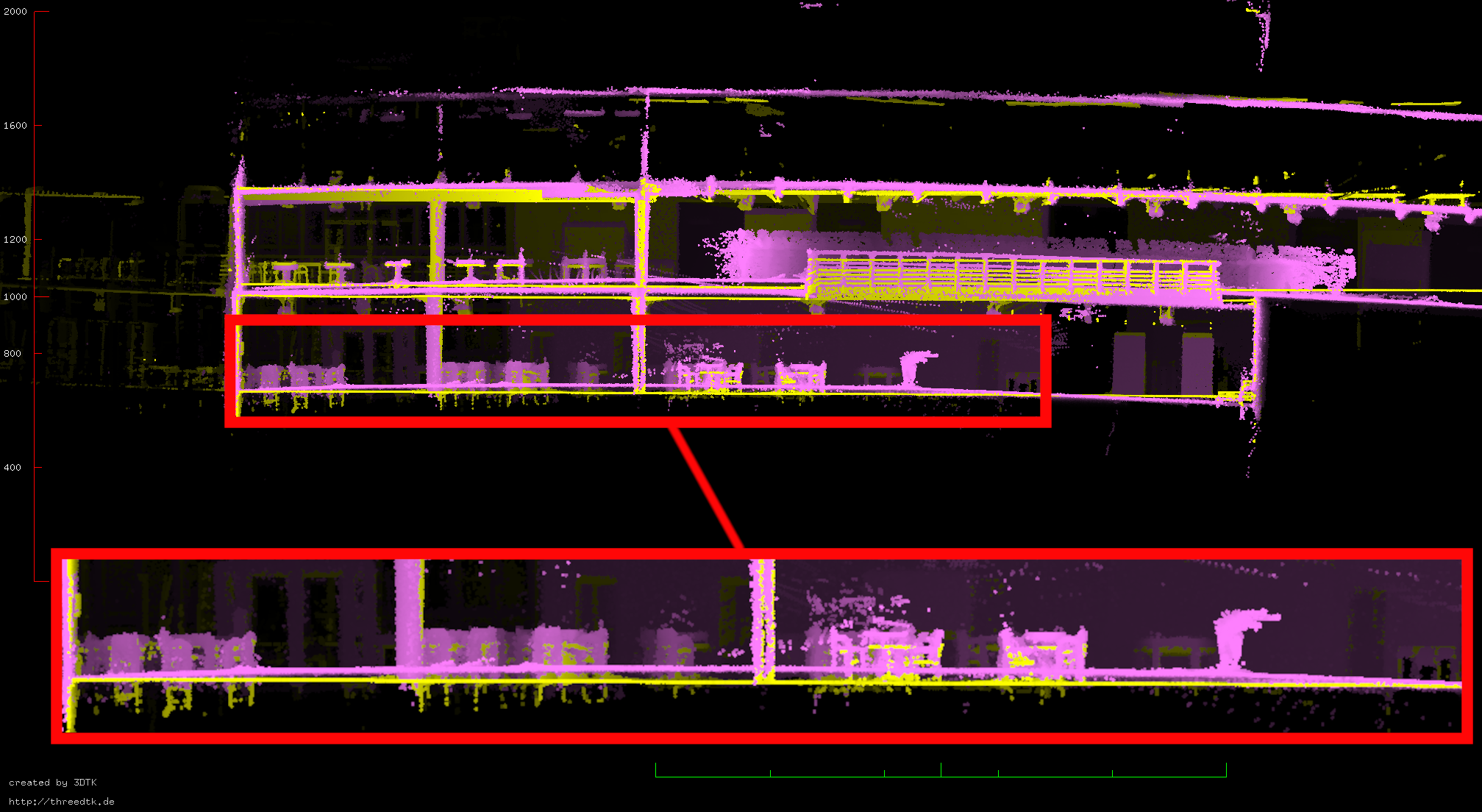}
    \caption{FAST-LIVO2 point-cloud}\label{fig:livo_bending}
\end{subfigure}
\caption{Cross section comparison. 
Yellow: Ground truth, Magenta: Mentioned algorithm.
The red boxes indicate noticeable bending of the ground plane.}\vspace{-1mm}
\label{fig:bending}
\end{figure}

The LIO algorithms exhibited small drift and bending over time, which is particularly noticeable in Fig.~\ref{fig:bending}. 
We were unable to obtain a satisfactory map using DLIO from the non-actuated sphere as shown in Fig.~\ref{fig:dlio_drift}.
Furthermore, sometimes the algorithms drifted without recovering after a fast motion, as shown in Fig.~\ref{fig:lio_drift}. 
We attribute this to the underlying motion models used in the LIO algorithms, which were designed for more conventional systems.

\begin{figure*}
\centering
\begin{subfigure}{0.195\textwidth}
    \centering
    \includegraphics[width=\textwidth]{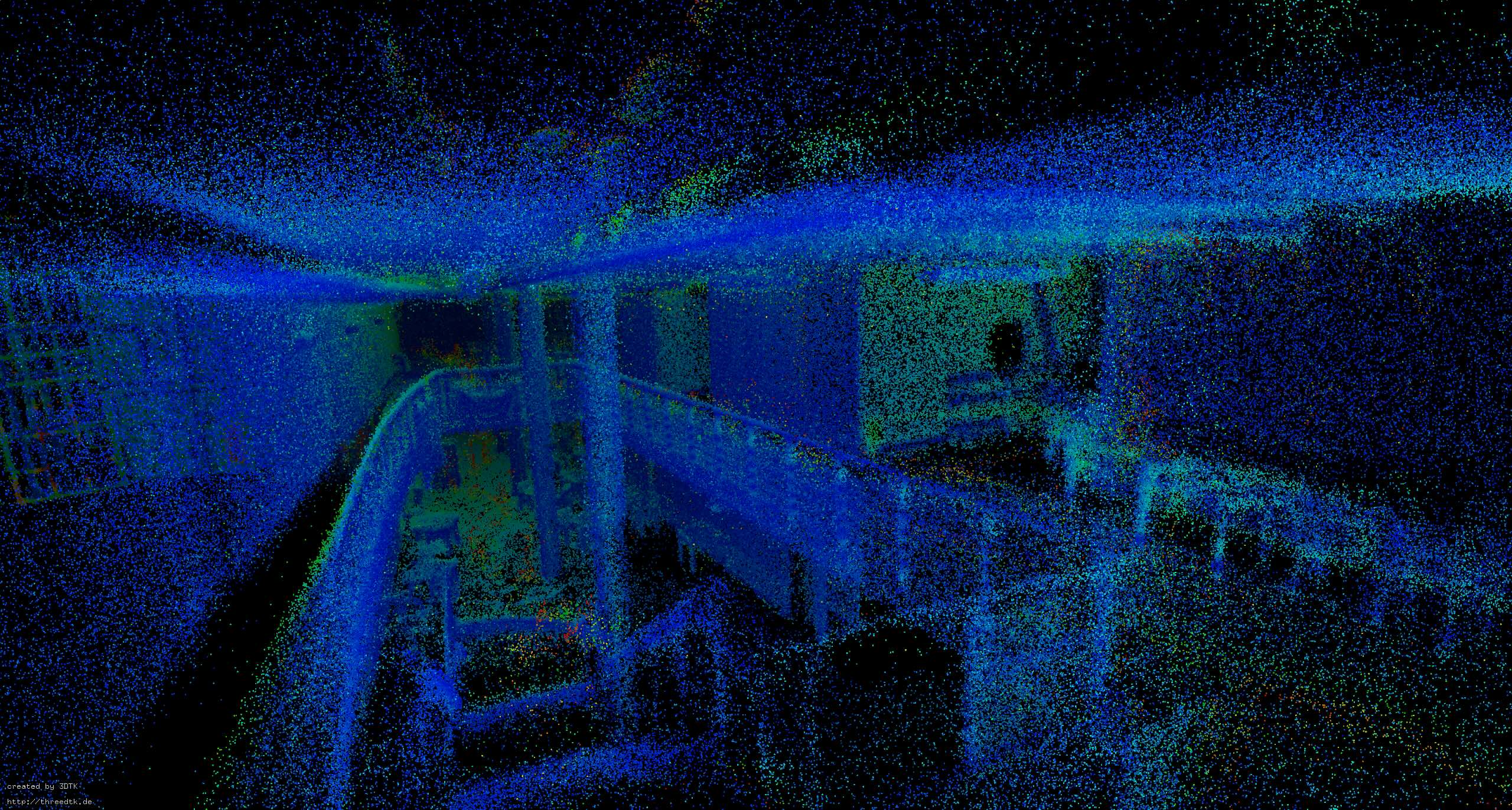}
    \label{fig:results_non_lio}
\end{subfigure}
\begin{subfigure}{0.195\textwidth}
    \centering
    \includegraphics[width=\textwidth]{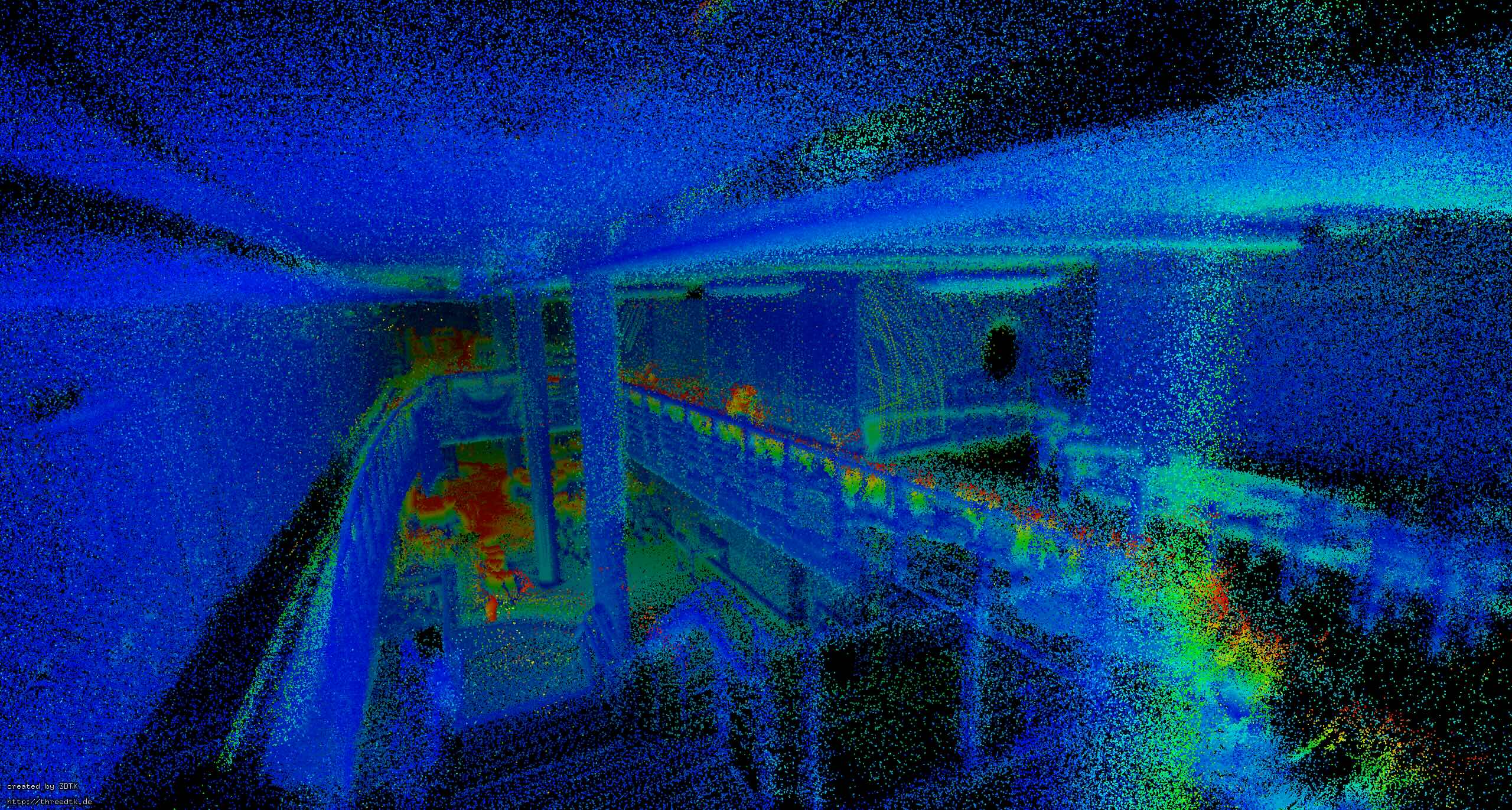}
    \label{fig:results_non_livo}
\end{subfigure}
\begin{subfigure}{0.195\textwidth}
    \centering
    \includegraphics[width=\textwidth]{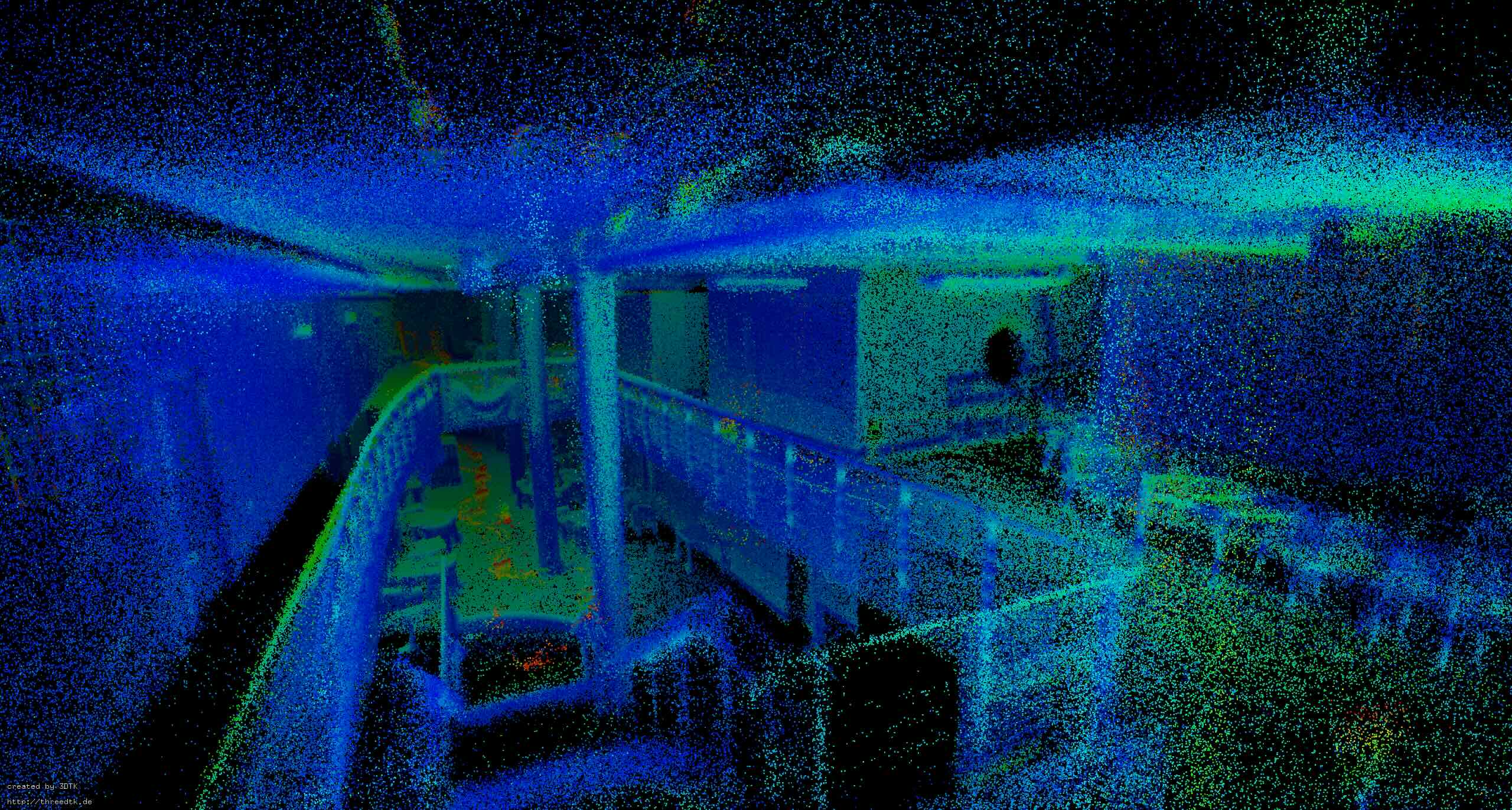}
    \label{fig:results_act_lio}
\end{subfigure}
\begin{subfigure}{0.195\textwidth}
    \centering
    \includegraphics[width=\textwidth]{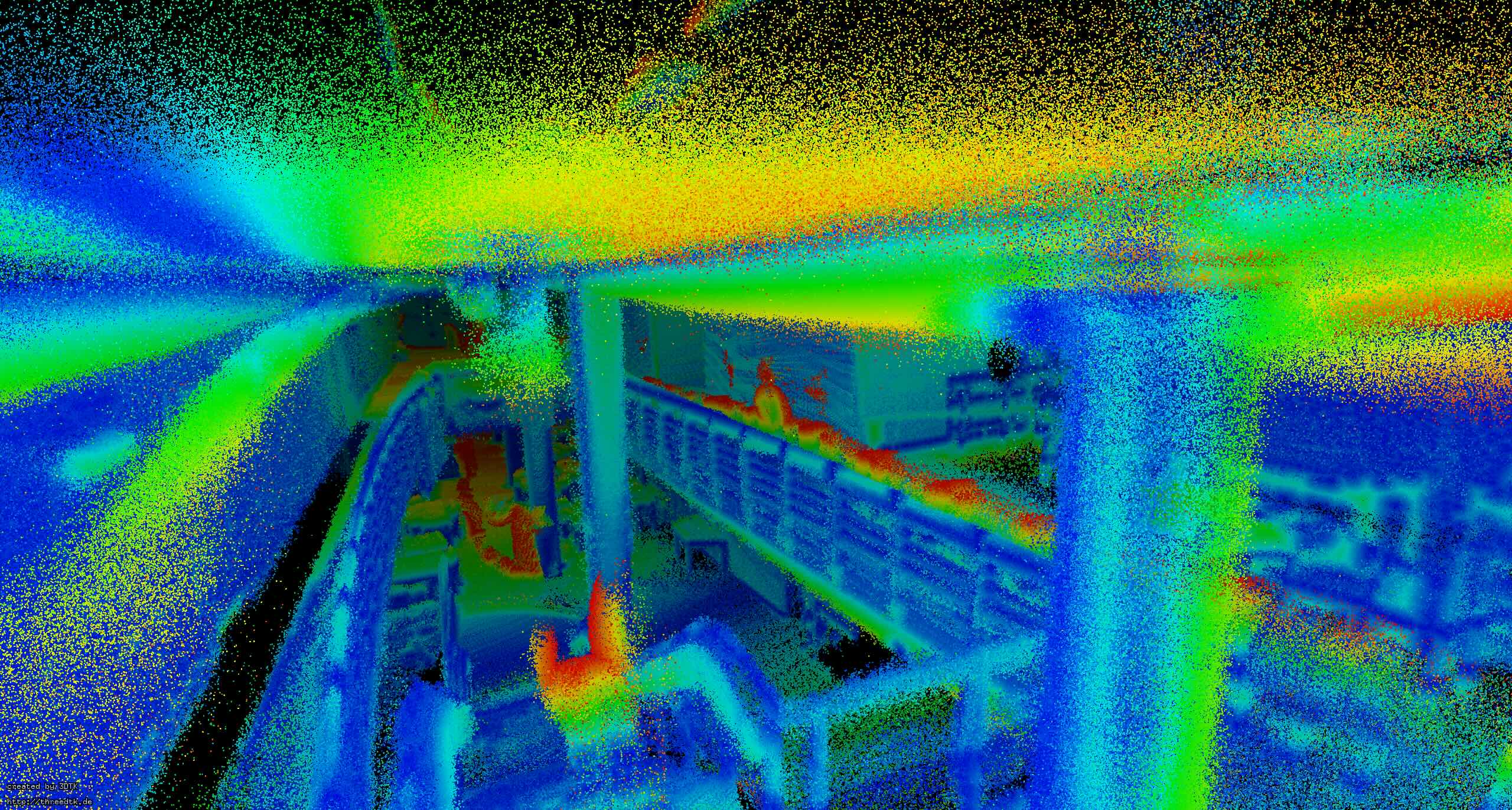}
    \label{fig:results_act_dlio}
\end{subfigure}
\begin{subfigure}{0.195\textwidth}
    \centering
    \includegraphics[width=\textwidth]{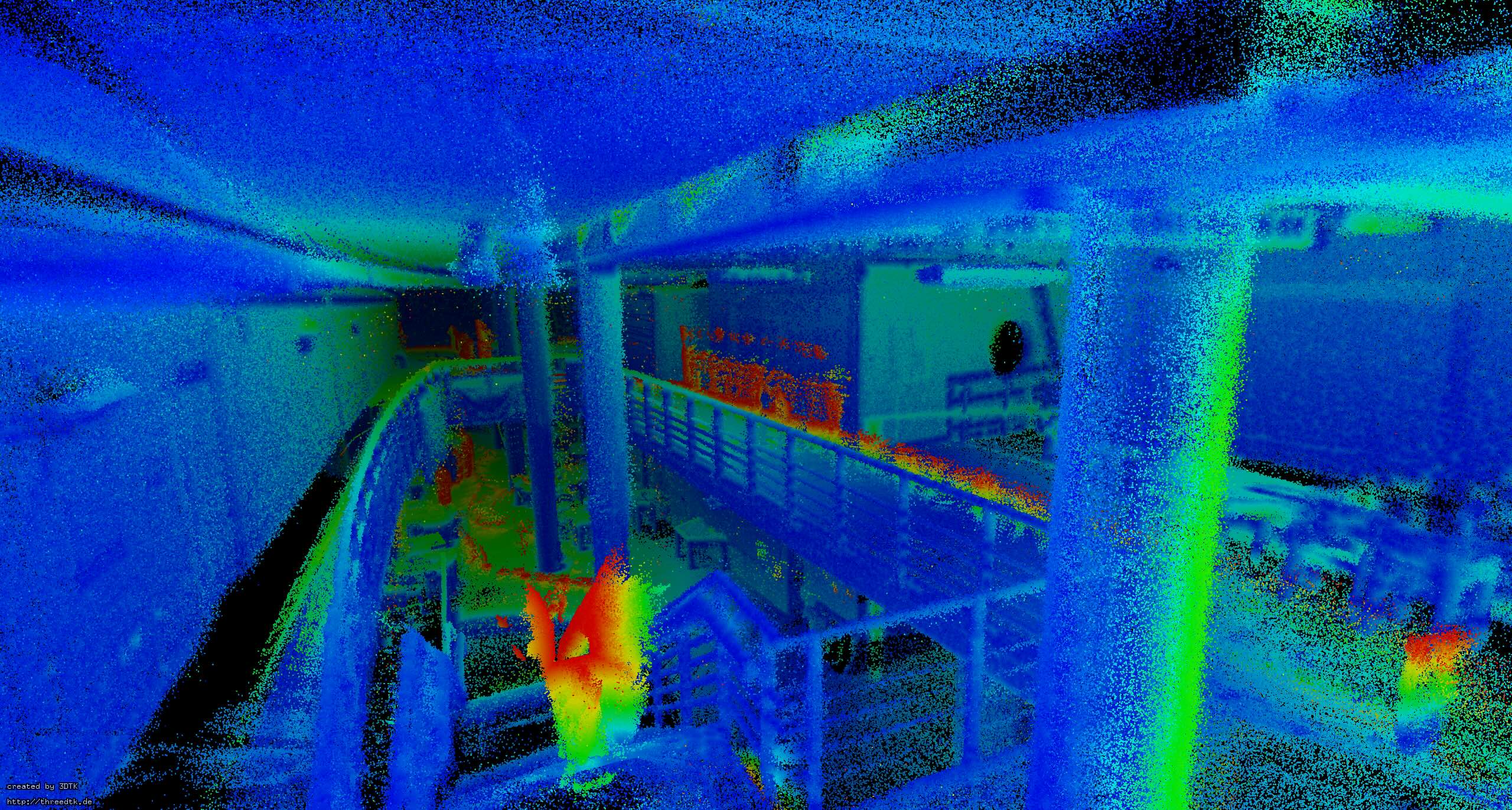}
    \label{fig:results_act_livo}
\end{subfigure}

\begin{subfigure}{0.19\textwidth}
    \centering
    \includegraphics[width=\textwidth]{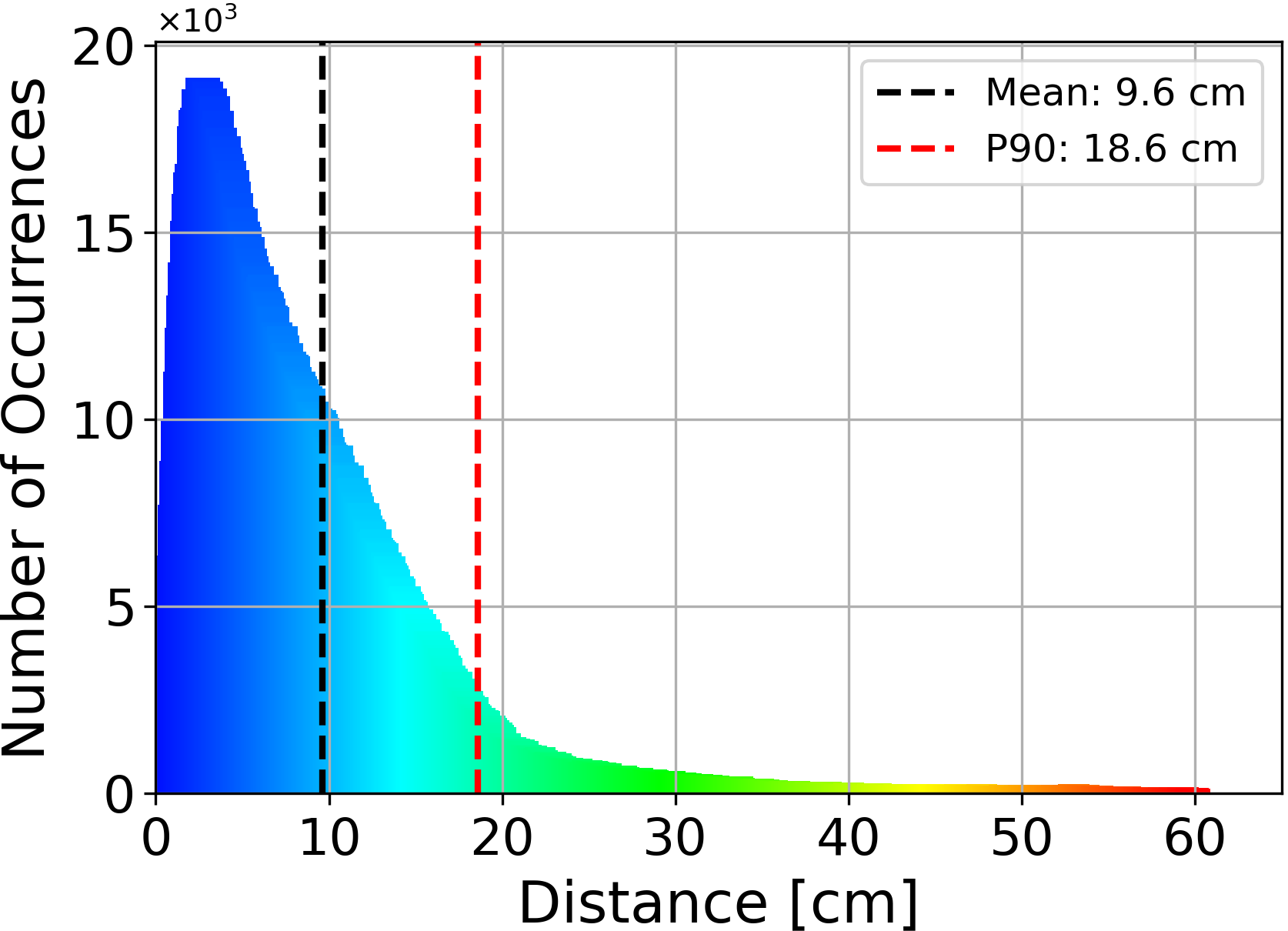}
    \caption{Non-Act. FAST-LIO2}
    \label{fig:hist_non_lio}
\end{subfigure}
\hfill
\begin{subfigure}{0.19\textwidth}
    \centering
    \includegraphics[width=\textwidth]{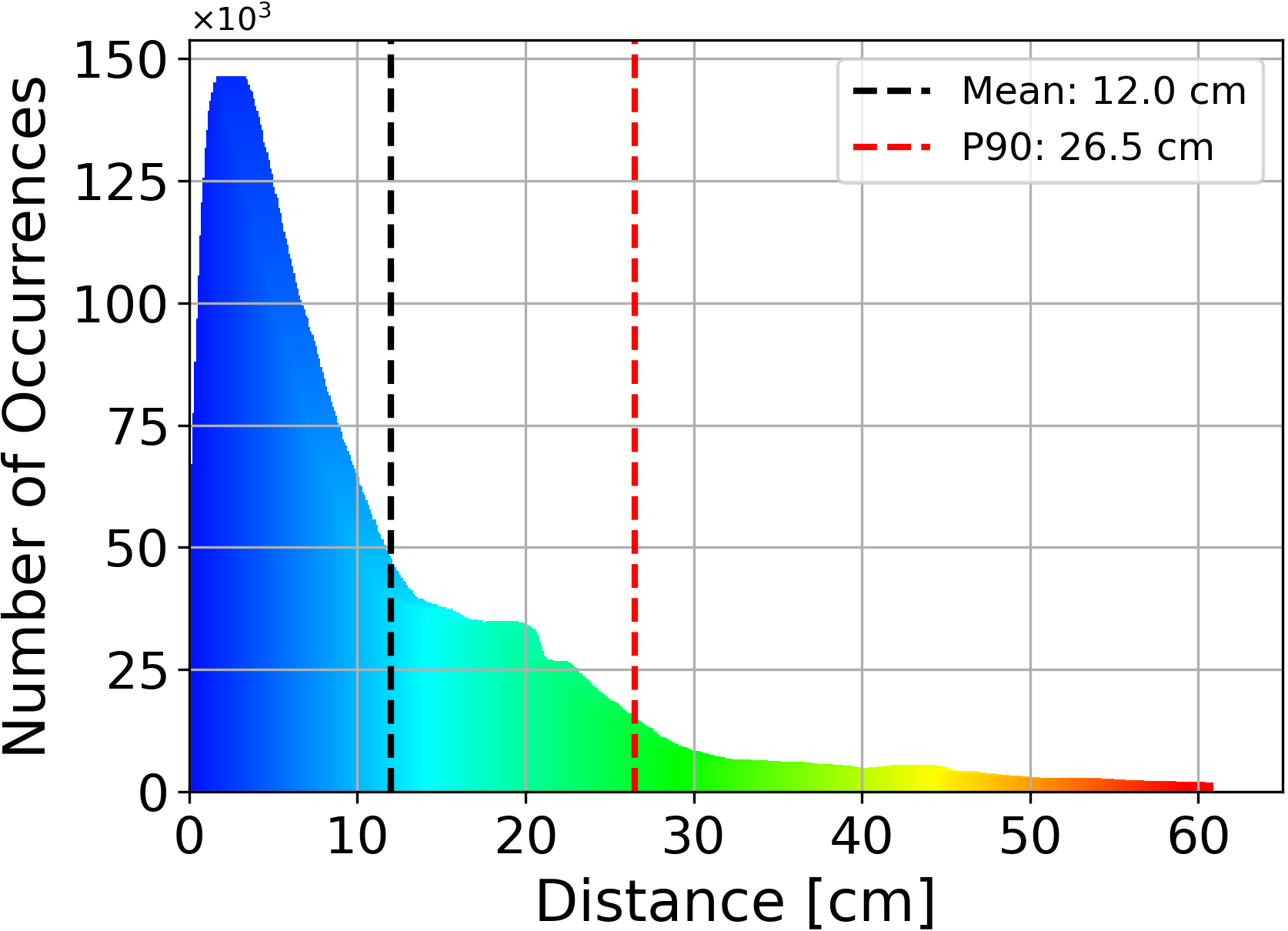}
    \caption{Non-Act. FAST-LIVO2}
    \label{fig:hist_non_livo}
\end{subfigure}
\hfill
\begin{subfigure}{0.19\textwidth}
    \centering
    \includegraphics[width=\textwidth]{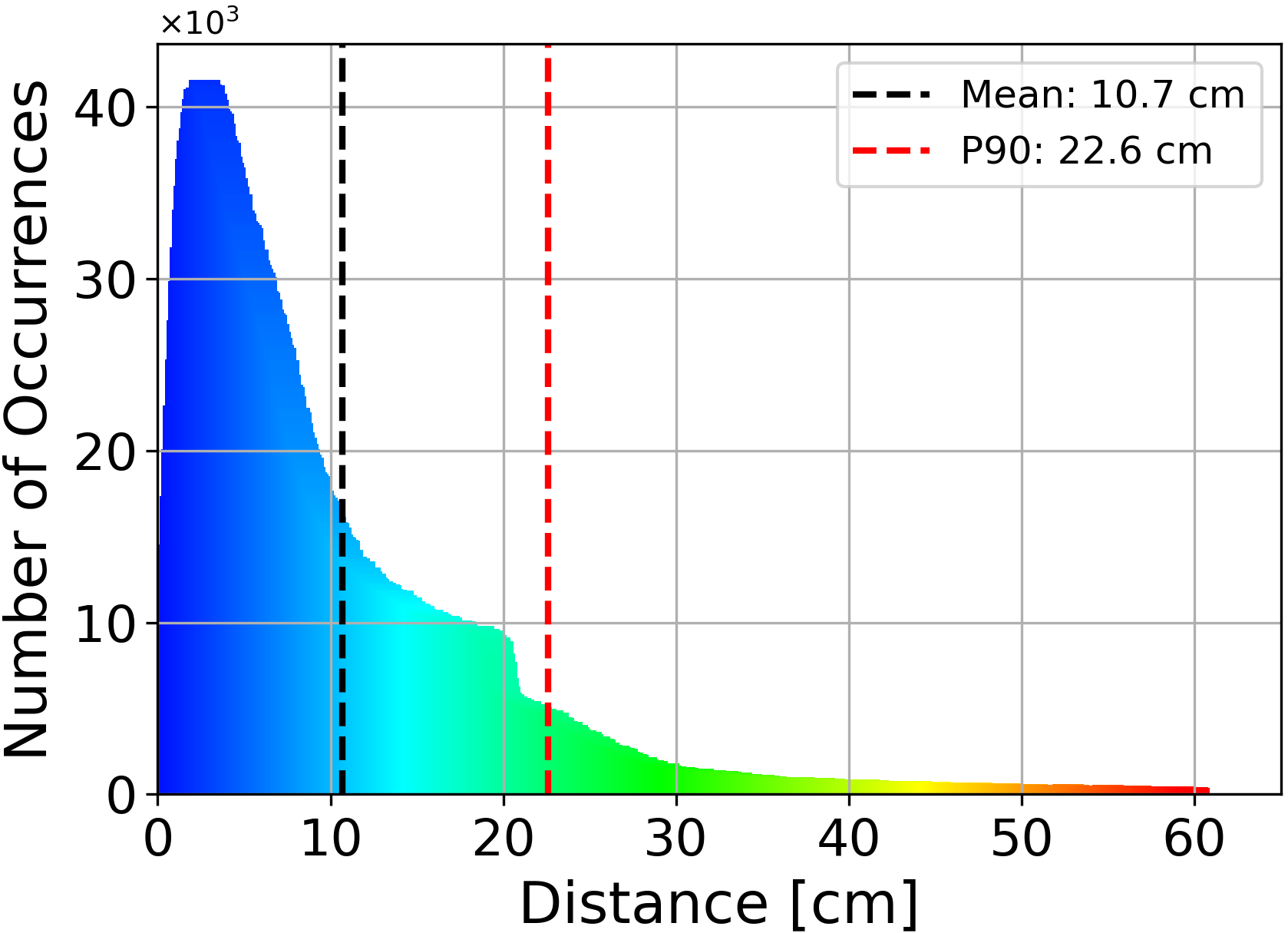}
    \caption{Act. FAST-LIO2}
    \label{fig:hist_act_lio}
\end{subfigure}
\hfill
\begin{subfigure}{0.19\textwidth}
    \centering
    \includegraphics[width=\textwidth]{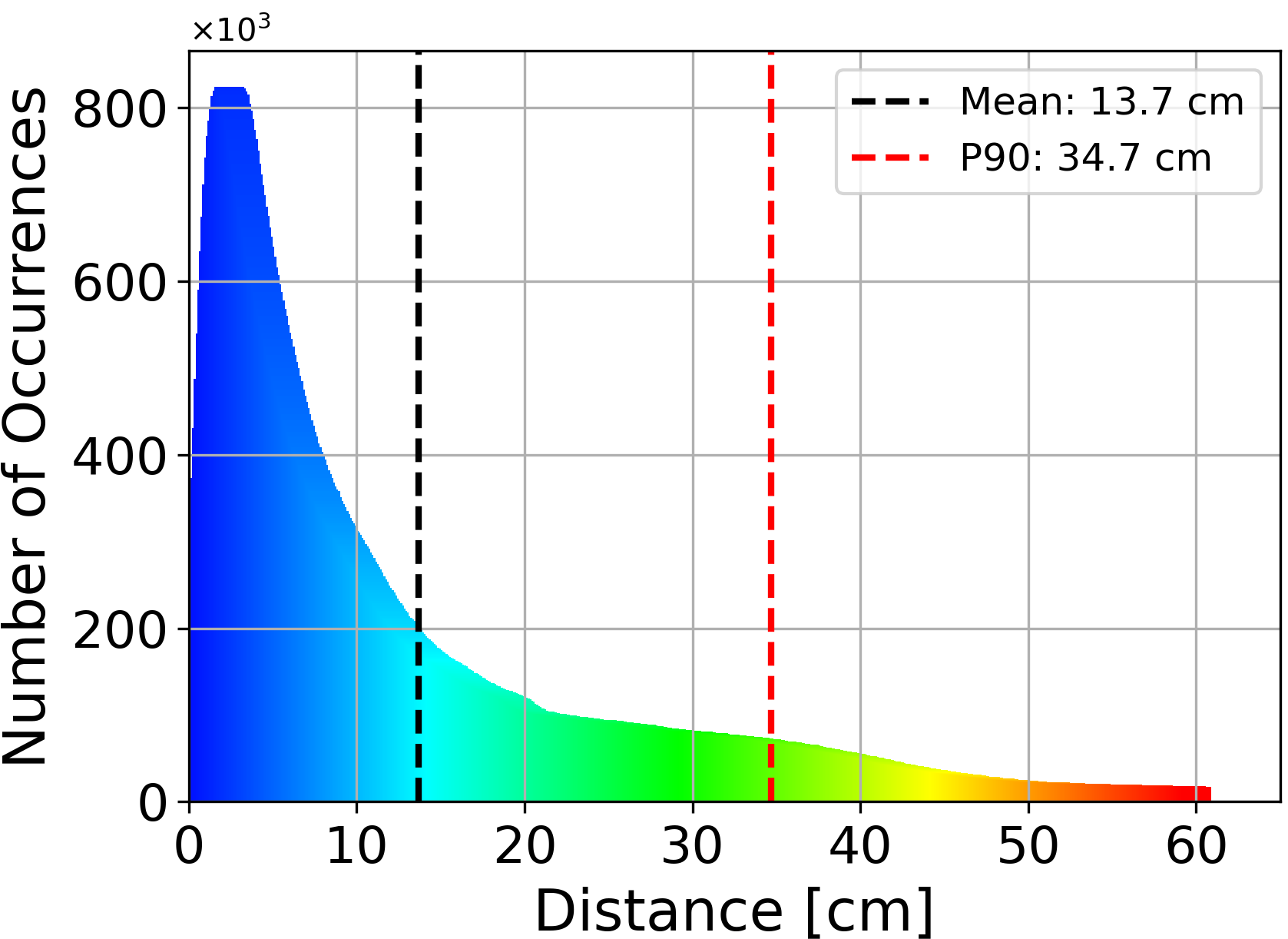}
    \caption{Act. DLIO}
    \label{fig:hist_act_dlio}
\end{subfigure}
\hfill
\begin{subfigure}{0.19\textwidth}
    \centering
    \includegraphics[width=\textwidth]{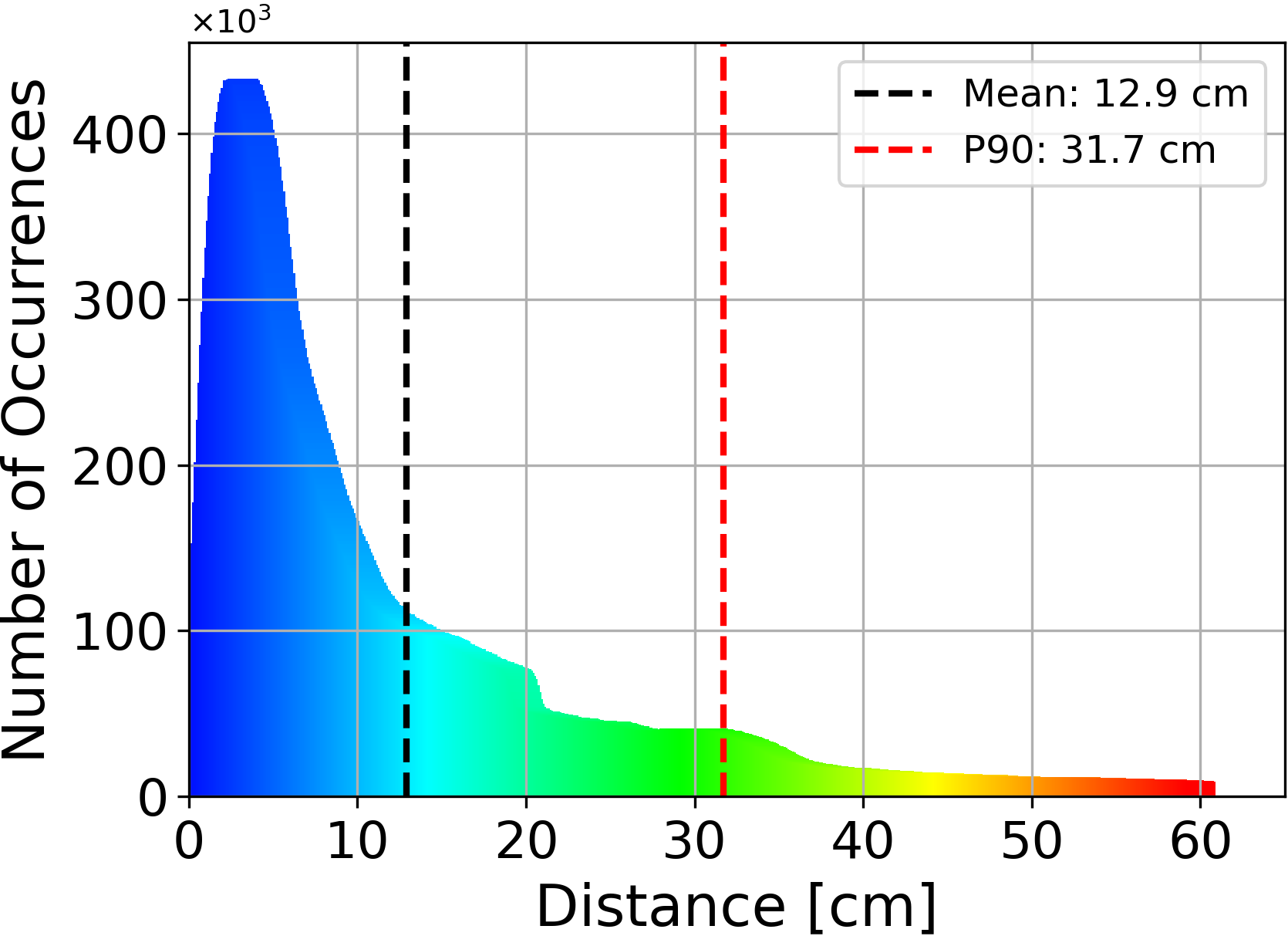}
    \caption{Act. FAST-LIVO2}
    \label{fig:hist_act_livo}
\end{subfigure}\vspace{2mm}
\begin{subfigure}{0.16\textwidth}
    \centering
    \includegraphics[width=\textwidth]{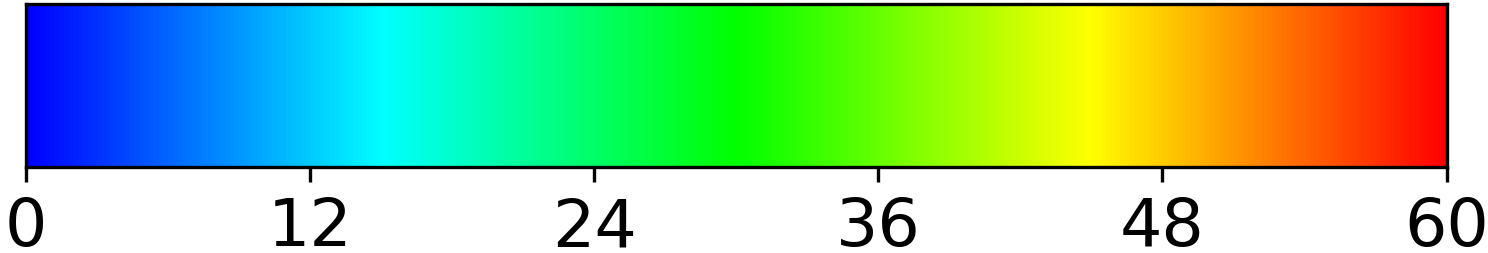}
    \caption{Color Mapping}
    \label{fig:hsv}
\end{subfigure}
\caption{Point-cloud Results and Error Distribution Analysis. A fly through video of the point-clouds is available at \url{https://youtu.be/Ere4UjPg-gk}. 
The first row shows the point-clouds generated by comparing the RIEGL map with each algorithm, while the second row displays the corresponding error distribution histograms.}\vspace{-2mm}
\label{fig:combined_results}
\end{figure*}

\section*{Conclusions}
In this paper, we presented the design and evaluation of two spherical robots for 3D mapping applications: a non-actuated, and a self-actuated sphere. 
The self-actuated sphere uses a pendulum-based locomotion mechanism, enabling controlled movement and stabilization in addition to mapping capabilities.
This is, to the best of our knowledge, the first prototype of a self-actuated spherical robot performing online LIO.   
The mapping accuracy of both systems was evaluated in a controlled indoor environment using ground truth point-clouds. 
The non-actuated sphere using FAST-LIO2 surprisingly achieved the lowest mean error and RMSE among all configurations.
We attribute this to the placement of the LiDAR sensor, which is closer to the center compared to the actuated sphere.
All tested algorithms produce bent point-clouds, and sometimes unrecoverable drift due to the rotationally aggressive motion of the system.  
In future work, we plan to incorporate motion models into the LIO algorithms that reflect the motion of the sphere better.
Furthermore, we want to explore alternative actuation mechanisms.
Overall, this work contributes to the growing field of spherical robots and provides a foundation for future advancements in autonomous 3D mapping and mobile perception.
\bibliographystyle{IEEEtran}
\bibliography{bibliography}

\vspace{12pt}

\end{document}